\documentclass[]{kling/kling}

\usepackage{microtype}
\usepackage{graphicx}
\usepackage{subcaption}
\usepackage{csquotes}
\usepackage{afterpage}
\usepackage{xcolor}

\usepackage{amsmath}
\usepackage{amssymb}
\usepackage{mathtools}
\usepackage{amsthm}
\usepackage{xspace}
\usepackage{colortbl}
\usepackage{multirow}
\usepackage{enumitem}
\usepackage{wrapfig}
\usepackage{nicefrac}
\usepackage[font=small,labelfont=bf,justification=centering]{caption}

\usepackage{booktabs}
\usepackage{multirow}
\usepackage{graphicx}
\usepackage{xcolor}
\usepackage{pifont} 

\usepackage{geometry}
\usepackage{tabularx}

\newcommand{\cmark}{\textcolor{green!60!black}{\ding{51}}}%
\newcommand{\xmark}{\textcolor{red}{\ding{55}}}%

\usepackage{amsmath}

\title{Kling-Omni Technical Report}
\headertext{Kling-Omni: Omni Video Generation Model from Multi-modal Visual Language}

\author[]{Kling Team, Kuaishou Technology}

\abstract{
We present \textbf{Kling-Omni}, a generalist generative framework designed to synthesize high-fidelity videos directly from multimodal visual language inputs. Adopting an end-to-end perspective, Kling-Omni bridges the functional separation among diverse video generation, editing, and intelligent reasoning tasks, integrating them into a holistic system. Unlike disjointed pipeline approaches, Kling-Omni supports a diverse range of user inputs, including text instructions, reference images, and video contexts, processing them into a unified multimodal representation to deliver cinematic-quality and highly-intelligent video content creation. To support these capabilities, we constructed a comprehensive data system that serves as the foundation for multimodal video creation. The framework is further empowered by efficient large-scale pre-training strategies and infrastructure optimizations for inference. Comprehensive evaluations reveal that Kling-Omni demonstrates exceptional capabilities in in-context generation, reasoning-based editing, and multimodal instruction following. Moving beyond a content creation tool, we believe Kling-Omni is a pivotal advancement toward multimodal world simulators capable of perceiving, reasoning, generating and interacting with the dynamic and complex worlds.
}

\metadata[Date]{December 18, 2025}
\metadata[Access]{\url{ https://app.klingai.com/global/omni/new}}
\metadata[Model ID]{Kling-Omni  (Kling-O1)}

\begin{document}

\maketitle

\section{Introduction}
\label{section:intro}
It has been a long-term vision in artificial general intelligence to create multimodal assistants capable of perceiving, reasoning, and creating across all sensory domains and generating visual outputs that mirror human communication through language~\cite{achiam2023gpt,team2023gemini,liu2024deepseek}, visual demonstration~\cite{rombach2022high,wu2025qwen}, and temporal dynamics~\cite{brooks2024video,runway_aleph}. Ideally, such systems should seamlessly process diverse inputs, whether text, images, or video, and produce corresponding outputs.

Recent breakthroughs in unified modeling~\cite{hurst2024gpt,ma2025janusflow} have brought this vision closer to reality. Pioneering works in image-text unification have successfully bridged the gap between understanding and generation by jointly optimizing these capabilities within a unified architecture. Models like Gemini 3 Pro Image~\cite{gemini3proimage} have further accelerated this paradigm shift, evolving from specialized single-task solvers into comprehensive systems that integrate computer vision, reasoning, and content creation. These advances signal a decisive move away from fragmented "expert models" toward powerful, general-purpose unified systems.

Despite this progress, integrating video understanding and generation remains a significant challenge due to the following reasons. \textit{Firstly, the current landscape of video generation~\cite{gao2025seedance,wan2025wan,veo3} is still dominated by fragmented approaches.} Most state-of-the-art video models are narrowly focused on specific tasks, such as text/image-to-video synthesis, and often rely on static text encoders that struggle to capture complex visual details. On the other hand, video editing and understanding frequently depend on separate, task-specific pipelines or external adapters, which complicates scaling and integration. As a result, advanced capabilities that require a deep synergy between perception and creation—such as multimodal in-context generation, precise visual editing through reasoning, and responding to interleaved video-text instructions—are still out of reach for existing video architectures. 
\textit{Secondly, the interaction paradigm towards a unified video generation system remains a significant bottleneck.} Relying solely on natural language prompts often fails to capture the nuances of visual imagination; text is inherently limited in describing precise spatial relationships, visual references, and temporal dynamics, leading to a gap between user intent and model output. \textit{Finally, current models~\cite{brooks2024video} lack deep, native intelligence.} While they excel at pixel-level synthesis, they often struggle with semantic reasoning and understanding the underlying physics or logic of a scene, acting more as passive generators than intelligent agents capable of inferring complex user intentions.

In this work, we introduce Kling-Omni, a generalist framework designed to tackle these challenges by unifying diverse video generation, editing, and intelligent creation tasks. Kling-Omni employs a straightforward architecture, representing an important step from specialized expert models to a unified system that seamlessly integrates these capabilities and removes task boundaries.

To achieve this, we propose multi-modal vision language (MVL) as a new interaction paradigm, revolutionizing how users interact with video generation models. Unlike traditional approaches, MVL constructs a unified input representation by combining natural language as a semantic skeleton with multi-modal descriptions. This enhances the model's foundational understanding and control by treating text and visual signals as a cohesive language. 

Moreover, Kling-Omni represents an advancement towards multi-model intelligence. The introduction of MVL does not merely refine instruction following; it empowers the model to deeply understand and infer user intentions. By exploring this inference potential, Kling-Omni moves beyond rote generation, demonstrating unexpected reasoning capabilities.

The remainder of this report is organized as follows. Sec. \ref{section:stage1} presents the methodology, introducing the key components, training strategies, training optimization, and inference optimization. Sec. \ref{section:data} focuses on data engineering, outlining the data collection and processing processes. Sec. \ref{section:performance} provides a comprehensive analysis of Kling-Omni's capabilities, including human evaluation, multi-modal referencing, interactive editing, and the broader potential of the model in intelligent reasoning and generation. Finally, we conclude with a discussion and acknowledges the contributions of the authors involved in the project.

\section{Methodology}
\label{section:stage1}

\subsection{Model Architecture Overview}
\begin{figure}[htbp]
    \centering\includegraphics[width=\textwidth]{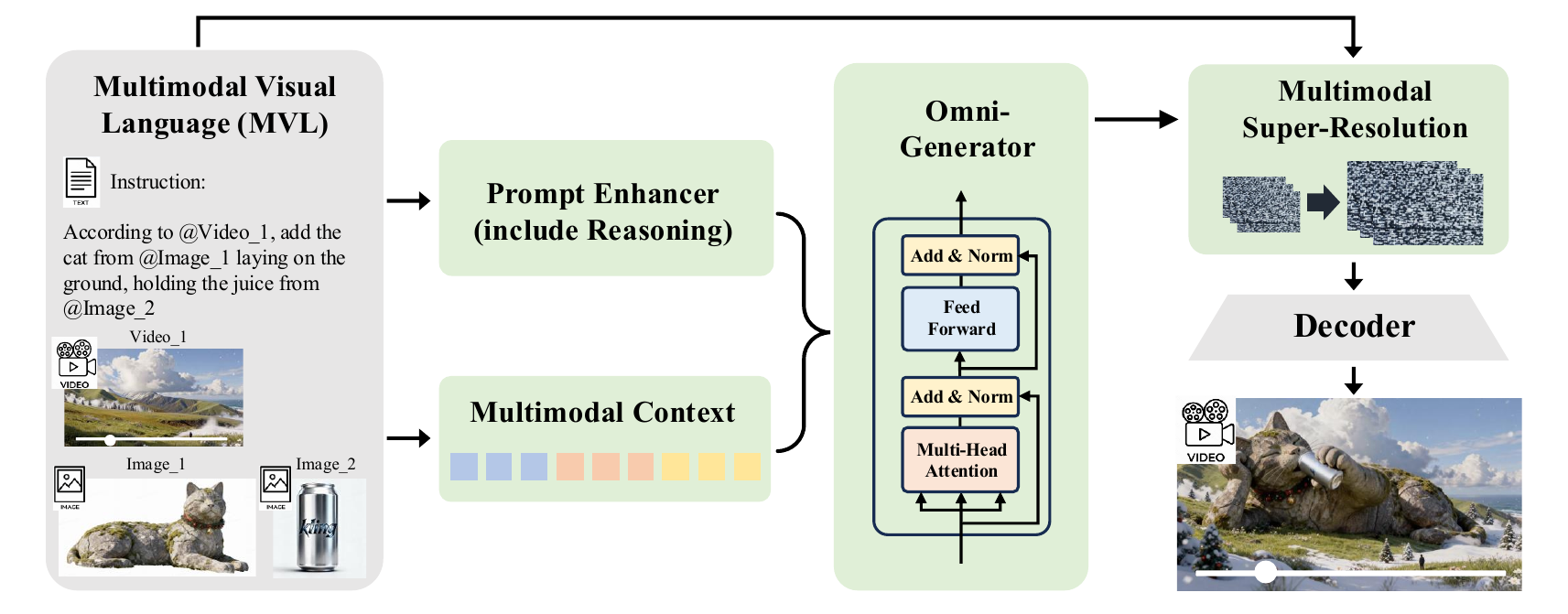} 
    \caption{Overview of Kling-Omni, a generalist framework that introduces multimodal visual language as the interaction mechanism, supporting diverse tasks including video generation, editing, and intelligent reasoning.}
    \label{fig:model_overview}
\end{figure}

We present \textbf{Kling-Omni}, a generalist generative framework designed to synthesize high-fidelity videos directly from multimodal visual language (MVL) inputs. Adopting an end-to-end perspective, Kling-Omni moves beyond disjointed pipeline approaches, integrating instruction understanding, visual generation and refinement into a holistic system. The architecture is designed to accept a diverse range of user inputs—including text instructions, reference images, and video contexts—processing them through a unified interface to produce cinematic-quality video content creation and editing with high intelligence.

As illustrated in the framework, the architecture comprises three key components underpinned by a robust training and infrastructure ecosystem. First, to bridge the gap between heterogeneous user inputs and the model's representations, a \textbf{Prompt Enhancer (PE)} module employs an MLLM to comprehend complex user inputs and synthesize them with learned world knowledge. By doing so, it infers the creator’s specific creative intent and reformulates the prompt accordingly. These refined features serve as input for the \textbf{Omni-Generator}, which processes visual and textual tokens within a shared embedding space, enabling deep cross-modal interaction, ensuring robust visual consistency and precise instruction adherence. 
The generated content is subsequently refined by a \textbf{Multimodal Super-Resolution} module, which conditions on original MVL signals to refine high-frequency details. The entire system is empowered by a progressive multi-stage training strategy, ranging from instruction pre-training, supervised fine-tuning to reinforcement learning (RL), and operates on a highly optimized infrastructure utilizing 3D parallelism and model distillation to improve training and inference efficiency.

\subsection{Training Strategies of Omni-Generator}
\subsubsection{Pre-training}
In the pre-training phase, we harness large-scale text-video paired data to instill robust instruction-based text-to-video generation capabilities into the model. To ensure the model can adapt to a wide spectrum of user inputs, we curate captions ranging from concise prompts to elaborate narratives, thereby laying a solid groundwork for comprehending diverse instructional formats. Furthermore, to catalyze the model's sensitivity to multi-modal vision-language (MVL) contexts, we infuse image-to-video tasks into the training mixture, establishing an early synergy between visual and textual modalities.

\subsubsection{Supervised Fine-tuning}
\textbf{Continue-training.} This stage focuses on deeply aligning the model with complex MVL inputs. We introduce a comprehensive curriculum that includes reference-to-video generation, image/video editing, and a suite of specialized tasks for semantic understanding. These tasks feature highly interleaved formats of image, video, and text conditioning. By exposing the model to such heterogeneous and information-rich data, we effectively enhance its ability to interpret intricate instructions and perform preliminary reasoning.

\textbf{Quality-tuning.}  To further enhance the generation quality and multimodal understanding capacity of the model, we have meticulously constructed a high-quality dataset characterized by a balanced task distribution and exceptional video standards. Each data sample is paired with precise instruction annotations. Through iterative fine-tuning on this premium dataset, we progressively optimize the model's output distribution, steering it towards a domain of superior visual quality and understanding capacity.

\subsubsection{Reinforcement Learning}
To bridge the gap between model outputs and human aesthetic preferences, we employ Direct Preference Optimization (DPO) \cite{rafailov2023direct}. We favor DPO over alternative algorithms like GRPO\cite{shao2024deepseekmath} because it bypasses the computationally expensive trajectory sampling required by the latter, offering a streamlined one-step diffusion forward process. 

Our optimization objectives are centered on key perceptual metrics, specifically motion dynamics and visual integrity. For data construction, we sample a diverse array of MVL conditions to form a candidate pool, subsequently generating multiple video variations using distinct random noise. These variations are then subjected to human evaluation to identify preference pairs—distinguishing between the optimal (preferred) and suboptimal (dispreferred) outcomes. During training, these preference pairs, along with their corresponding noise and timesteps, are utilized to compute the DPO loss. Through multiple rounds of this preference-aligned training, the model achieves significant improvements in video generation quality, aligning more closely with human intent. 

\subsubsection{Model Acceleration (Distillation)}

We develop a two-stage distillation methodology to substantially reduce the computational cost of inference while preserving output fidelity. 
The acceleration pipeline incorporates both trajectory matching distillation and distribution matching distillation, compressing the model inference to 10 Number of Function Evaluations (NFE), which originally costs 150 NFE to synthesize a single video sample before distillation.

In the first stage, the procedure follows the principle of trajectory matching distillation—exemplified by PCM \cite{wang2024phased}, HyperSD \cite{ren2024hyper}, and related methods—to ensure a closer adherence to the teacher trajectory at the early training phase.
Specifically, we employ a phase-wise temporal structuring of the training objective with the timestep scheduler partitioned into several phases. 
The student model is supposed to predict temporally consistent denoising outputs that align with the designated phase endpoint at any reverse step. 
Different from common practice that initially distills a student model into an intermediate state whose NFE is reduced yet still exceeds the expected NFE, we directly make the student execute with the target scheduler of 10 sampling step in this stage.

To further enhance the generation performance, distribution matching distillation is conducted as the second stage training. 
Unlike other score-based distillation algorithms such as DMD\cite{yin2024improved} and SiD\cite{zhou2024score} that formulate the student as a stochastic differential equation (SDE) process, we adopt the insights of TDM \cite{luo2025learning} and distill the student to perform few-step ordinary differential equation (ODE) sampling, which has been empirically demonstrated to be more suitable for our tasks. 
In addition, the trajectory matching objective is preserved at this stage, serving as a "regularization" mechanism to prevent the model from deviating significantly from the reference trajectory.
The similar operation has also been reported in \cite{cheng2025structure}.

\subsection{Prompt Enhancer}
\label{subsec:pe}
To address the ambiguity and high variance inherent in user inputs, we introduce a Prompt Enhancer (PE) module for Kling-Omni. The primary function of the PE is to map diverse user prompts onto a distribution that is consistent with the model's training data. This alignment is critical for enhancing generative quality, specifically in terms of identity preservation, spatial coherence, and color fidelity, while simultaneously improving physical plausibility via textual reasoning~\cite{wiedemer2025video,tong2025thinking}.

The PE is built upon a Multimodal Large Language Model (MLLM) to accommodate multi-modal user inputs. Since general-purpose MLLMs are not optimized for our specific generation tasks, we constructed a specialized multilingual dataset.

Our training pipeline involves two phases: initially, we utilize Supervised Fine-Tuning (SFT) to enable the model's reasoning chain (or "thinking process"). This is followed by Reinforcement Learning (RL), where the reward function is designed to maximize factual correctness, content richness, and semantic plausibility, as well as the similarity between the processed prompts and our high-quality training data. Experiments indicate that the PE module significantly boosts Kling-Omni's performance, resulting in videos with greater dynamism and detail. Furthermore, the PE demonstrates strong generalization potential, empowering the model with intelligent creativity.

\subsection{Multimodal Super-Resolution}
\label{subsec:sr}
\begin{figure}
    \centering
    \includegraphics[width=0.85\linewidth]{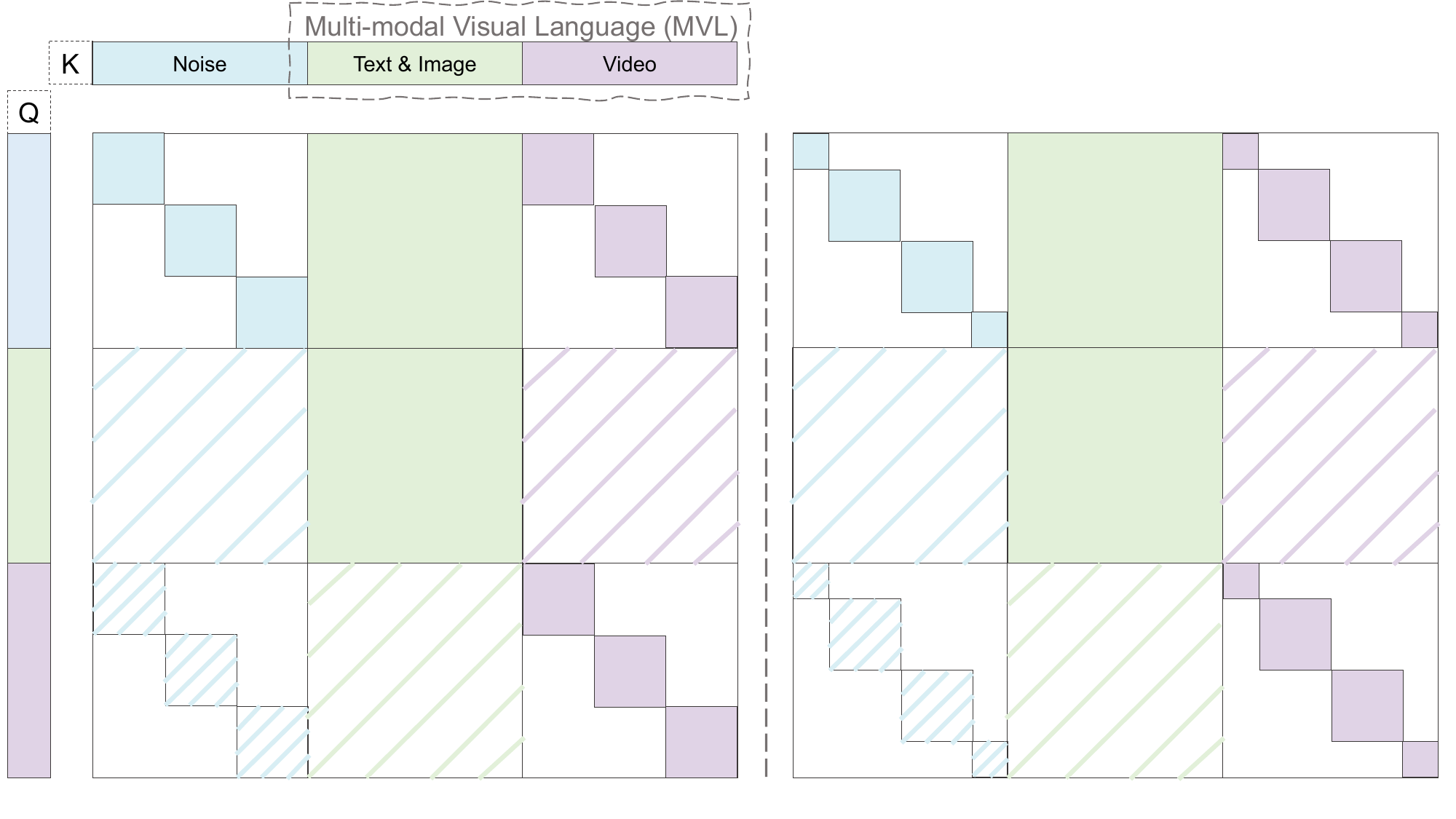}
    \caption{Attention maps in Multimodal Super-Resolution. The left panel illustrates the map for even-numbered layers, while the right panel shows the map for odd-numbered layers. Skipping the computation for the shaded regions leads to a substantial reduction in computational load and supports accelerated inference with a KV cache.}
    \label{fig:sr_attn}
\end{figure}
To improve the training and inference efficiency of generator, we propose a cascaded diffusion framework for Video Super-Resolution (VSR). Conditioned on both Low-Resolution (LR) latents from the base model and Multi-modal Vision-Language (MVL) signals, our VSR model operates as a unified framework. This cohesive design enables the synthesis of high-fidelity, fine-grained visual details and textures, catering to a diverse range of applications.

We adopt the architecture of the base model and initialize our VSR module using its pre-trained weights. To address the computational overhead imposed by long temporal contexts and high-resolution inputs, we exploit the inherent spatio-temporal redundancy of video data. Specifically, we replace standard full attention mechanisms with local window attention. To prevent receptive field isolation, we implement a shifted window strategy in every odd-numbered layer, offsetting the window by half its size, to facilitate information flow between adjacent non-overlapping windows, as illustrated in Fig. \ref{fig:sr_attn}.

To further minimize inference latency, we introduce an asymmetric attention mechanism. In this configuration, condition tokens (serving as queries) are restricted to self-attention, whereas noisy tokens attend to the full sequence. This decoupling allows us to cache the Key-Value (KV) features of the condition tokens, enabling their reuse across subsequent sampling steps. This strategy boosts generation efficiency with negligible impact on visual performance.

\subsection{Training Optimization} 

We develop an end-to-end training system that optimizes multimodal data processing, parallel execution, and computation kernels for large-scale pre-training.

\subsubsection{Multimodal Data Pipeline and Load Balancing}

\begin{figure}
    \centering
    \includegraphics[width=0.95\linewidth]{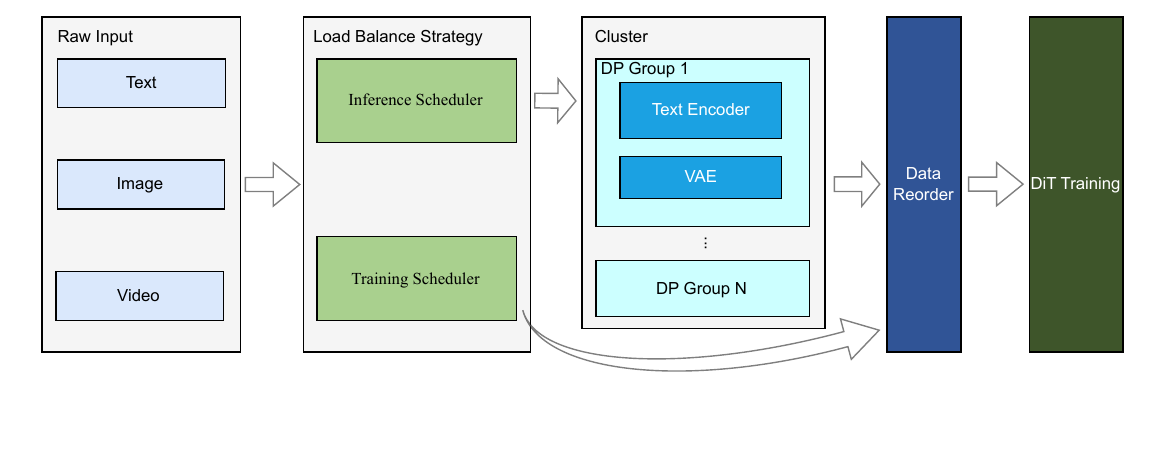}
    \vspace{-40pt}
    \caption{Online training data pipeline. Raw data is distributed across DP/PP groups using an inference scheduler. After inference, a training scheduler reorders data for balanced workload.}
    \label{fig:data_pipeline_flowchart}
\end{figure}

To handle significant sequence length variation across text, image, and video data, we employ a heuristic scheduling strategy to reduce imbalance bubble across pipeline-parallel (PP) \cite{fan2021dapple, huang2019gpipe, huang2024re, harlap2018pipedream, narayanan2021memory, narayanan2021efficient} and data-parallel (DP) groups. As shown in Fig.~\ref{fig:data_pipeline_flowchart}, the training loop is divided into two stages: online VAE/text encoder inference and DiT training. A central scheduler assigns samples to DP groups to ensure balanced workloads. For VAE/text encoder inference, tokens are dynamically partitioned across PP stages to balance encoding workloads and improve utilization.

To further address dynamic sequence lengths, we introduce a microbatch-level elastic ulysses-parallel (UP) \cite{jacobs2023deepspeed,liu2023ring} switching mechanism\cite{FlexSP}, as shown in Fig.\ref{fig:pipeline_schedule}. An online adaptive scheduler with asynchronous pipeline predetermines the UP degree per microbatch and dynamically adjusts assignments to DP ranks, reducing load imbalance. To mitigate network congestion from cross-node all-to-all communication, we adopt a two-tier all-to-all strategy (intra-node aggregation followed by inter-node exchange) to distribute traffic and alleviate spine switch workload.

\subsubsection{Efficient Multimodal Framework and Activation Reduction}

In DiT training, inputs are flattened into 1D sequences with minimal padding\cite{dehghani2023patch}, and the computation graph is restructured to preserve modality-independent computation, minimizing redundant data movement and layout transform overhead. A packing version of multimodal FlashAttention \cite{shah2024flashattention} operator (MM-FlashAttention) is developed to support arbitrary cross-modal masks and variable-length sequences within a single kernel while maintaining high performance.

For activation reduction, we selectively recompute \cite{MegatronKwai} the most cost-effective operators, and pipeline-aware offloading \cite{MegatronKwai} further reduces GPU memory by moving activations to CPU. Kernel fusions cut down memory traffic and launch overhead, which is crucial for packing phase. A virtual-pipeline-stage-aware mechanism reuses activations across model chunks with identical inputs, slashing memory and computation in multi-view, multi-stream scenarios.

\subsubsection{Reliability and High-Availability}

We achieve a 97\% Effective Training Time Ratio by compressing recovery time. An automated fault detection system monitors RDMA traffic to detect hangs within a minute, reducing worst-case exit time to minute-level. A custom TCP synchronization layer and concurrent artifact loading from NVMe enable sub-minute restarts. Parallelized warmup overlaps NCCL initialization and kernel compilation with I/O, reducing first-iteration overhead to second-level.

To ensure runtime stability, I/O operations are optimized through request-training overlap. Random reads from dataset shuffling are converted to sequential access via pre-shuffled Parquet files. Non-blocking asynchronous checkpointing and hardware isolation prevent interference. A unified observability stack correlates MFU drops with data shifts and kernel stalls for automated root-cause analysis.

\begin{figure}
    \centering
    \includegraphics[width=\linewidth]{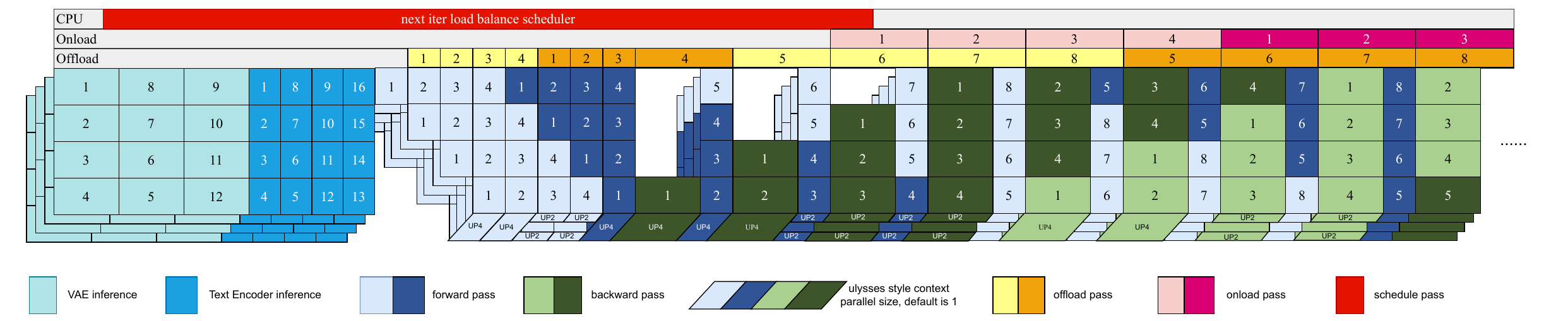}
    \vspace{-20pt}
    \caption{The pipeline schedule in Kling-Omni. The inference pass of VAE/TE are distributed across both data- and pipeline-parallelism, following an interleaved 1F1B pipeline schedule. Pipeine-aware offloading and onloading are introduced to reduce GPU memory consumption without blocking forward or backward pass, and an online  load balance scheduler is running on CPU to determine the ulysses parallel size and the workload for each microbatch.}
    \label{fig:pipeline_schedule}
    \label{fig:pipeline_schedule}
\end{figure}

\subsection{Inference Optimization} 

\textbf{Model Parallelism.} To mitigate the substantial GPU memory consumption and inference latency associated with long-sequence video generation, we adopt a hybrid parallel inference strategy, including Ulysses parallelism\cite{jacobs2023deepspeedulysses}, and tensor parallelism\cite{shoeybi2020megatronlm}. In addition, to reduce communication overhead, we design a computation–communication overlap scheme, which can hide most of the communication cost and has almost no impact on computation. 

\textbf{Quantization.} To further reduce inference latency and lower memory usage, we designed a comprehensive hybrid quantization scheme that achieves nearly lossless acceleration. The scheme has three main features:
    \begin{itemize}
        \item Wide quantization coverage. Most GEMM operations and self-attention modules in the model are quantized to FP8.
        \item Zero-overhead quantization. All quantization and dequantization operators are fused into other kernels, minimizing the additional overhead introduced by quantization
        \item FP8 communication. Using FP8 for communication further reduces communication overhead. When combined with communication-overlap techniques, most communication overhead can be effectively hidden.
    \end{itemize}

\textbf{Cache.} The Kling-Omni model takes a large number of reference images and reference videos as input, and these long conditional inputs significantly increase inference time. We designed a cache scheme tailored for Kling-Omni, achieving roughly a 2$\times$ speedup. In addition, we developed a cache-offload solution that greatly alleviates the potential memory pressure introduced by the caching mechanism. 

\section{Data System}
\label{section:data}
This section delineates the data methodology underlying our unified video generation and editing framework, structured around two pivotal dimensions: \textbf{data collection} and \textbf{data processing system}.

\begin{figure}[tbh]
    \centering
    \includegraphics[width=\textwidth]{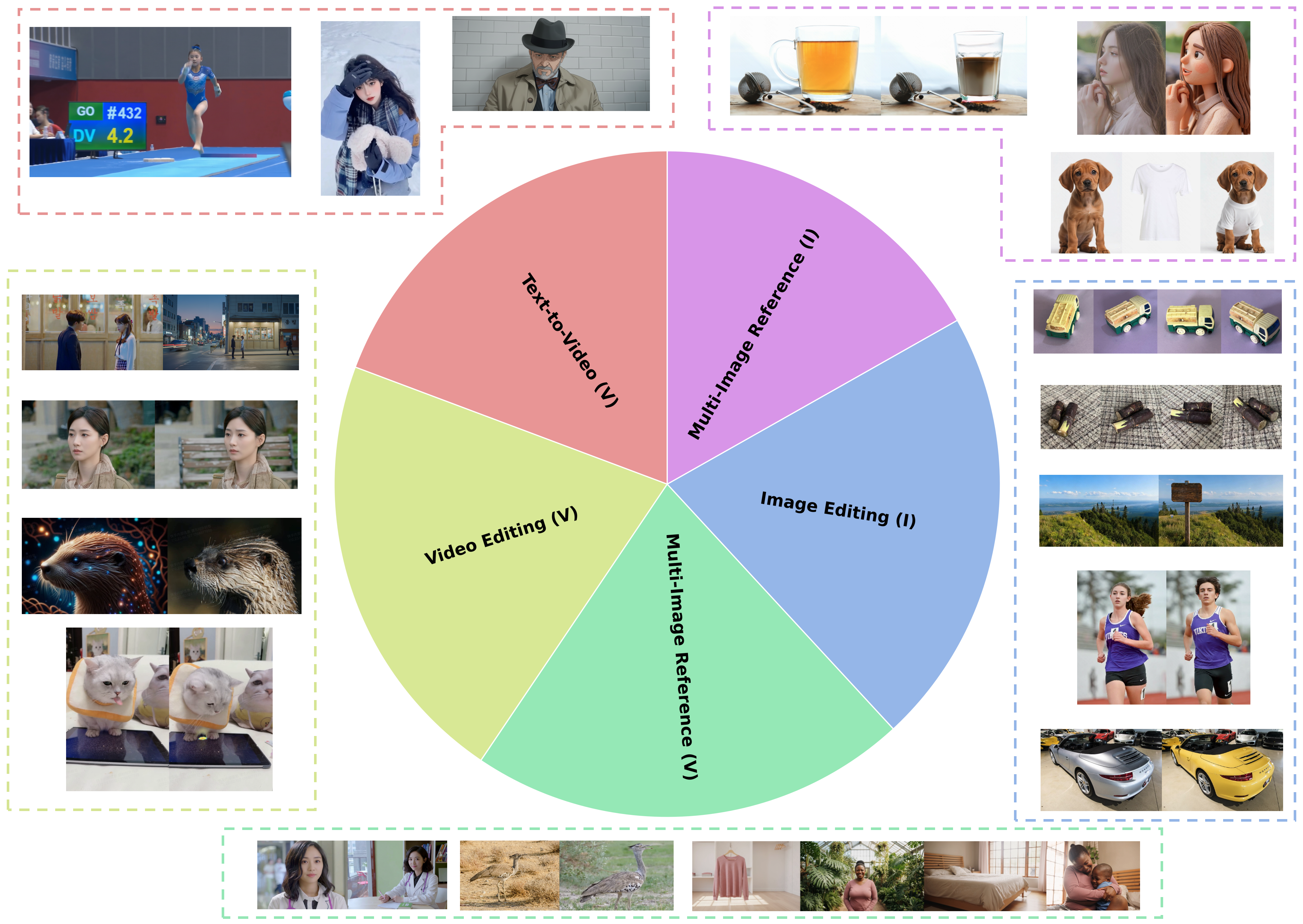} 
    \caption{Cross-modal and cross-task data distribution in our constructed data system.}
    \label{fig:data_distribution}
\end{figure}
Driven by the requirements of high-fidelity video synthesis, specifically regarding temporal consistency, semantic stability, multi-image reference alignment, and complex editing constraints, we have engineered a holistic data infrastructure. This system spans two key dimensions: cross-modality (image/text/video) and cross-task (image-to-video, video-to-video, editing, and reference-based generation, etc.), ensuring a robust foundation for model training, as illustrated in Fig.~\ref{fig:data_distribution}.

\subsection{Data Collection}

To construct a training corpus characterized by high diversity, consistency, and controllability, our data collection system integrates large-scale real-world data acquisition with task-oriented synthetic data construction.

\textbf{Real-World Data Acquisition.}
We curated a comprehensive collection of videos and images data to ensure broad scenario coverage. These sources provide essential natural priors, spanning diverse subjects, complex scenes, and stylistic variations. To expand task coverage beyond static datasets, we developed an automated pipeline for large-scale internet data mining. Utilizing an in-house embedding model, the pipeline identifies and constructs cross-modal samples that are semantically related or subject-consistent to enhance the model's generalization across diverse generation scenarios.

\textbf{Synthetic Data Construction.}
Since relying solely on real-world data is insufficient for learning precise controllability, we employed a synthesis pipeline driven by expert models. We utilized in-house image editing and video understanding models to produce high-quality samples for tasks such as editing and multi-image referencing. Furthermore, to support high-fidelity video generation tasks, we constructed an automatic reverse synthesis strategy. These approaches constructs robust reference-to-video training examples that preserve the temporal consistency of natural videos while providing explicit control signals.

\subsection{Data Processing}
\begin{figure}[t]
    \centering
    \includegraphics[width=\textwidth]{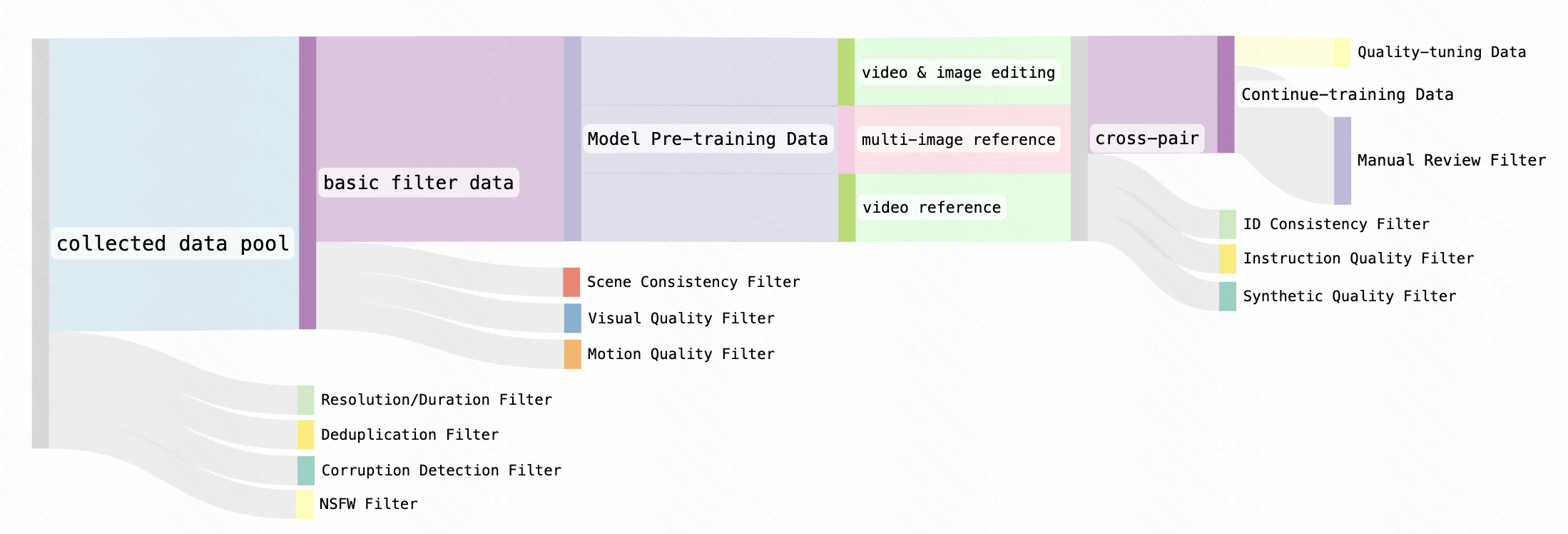} 
    \caption{Data filtering pipeline for video and image samples, illustrating the stages of quality control, temporal consistency, and multimodal alignment.}
    \label{fig:data_filter_pipeline}
\end{figure}

In large-scale multimodal training, data quality directly dictates the model's temporal consistency, semantic stability, and cross-modal alignment capabilities. We constructed a three-tier processing system covering basic governance, temporal stability, and cross-modal alignment to ensure that the training data exhibits a stable, clean, and interpretable distribution, as illustrated in Fig.~\ref{fig:data_filter_pipeline}.

\textbf{Basic Filtering.}
To establish a robust quality baseline, we implemented a rigorous governance protocol that filters unusable or compromised samples. This process begins with strict resolution and duration thresholds to ensure visual validity, followed by a deduplication mechanism using frame-wise and temporal fingerprinting to prevent model bias from redundant content. Additionally, we applied audio-visual corruption detection to eliminate samples with structural errors and enforce content safety protocols to exclude NSFW material. This foundational layer guarantees the hygiene of the raw data pool, preventing the training process from being disrupted by noise.

\textbf{Temporal Quality Assessment.}
Given the critical importance of temporal continuity in video generation, we employed specialized screening mechanisms for visual and temporal stability. We utilized quality scoring metrics to identify and penalize artifacts such as blur, jitter, and compression noise. To prevent the model from learning unnatural discontinuities, the system detects and removes abrupt scene changes and incoherent shot transitions. Furthermore, we filtered out videos with excessively low action semantic density, thereby improving the effective training ratio for dynamic content and ensuring the model learns high-quality temporal coherence.

\textbf{Video–Text and Image–Video Alignment.}
To support the unified modeling of text, images and videos, we established a systematic cross-modal alignment detection mechanism. This involves evaluating the semantic consistency between video captions and actual visual content, as well as assessing the fidelity of reference images to target videos for generation tasks. We further verified the alignment between editing instructions and their execution results. Crucially, for human-centric tasks, we enforced strict character identity consistency checks. These strategies ensure the model learns accurate mapping relationships across modalities, facilitating robust performance in complex editing and generation scenarios.
\section{Model Performance}
\label{section:performance}

This section presents a comprehensive evaluation and capability analysis of Kling-Omni. Specifically, Sec.~\ref{sec:human_eval} details the internal evaluation protocol, including the overall benchmark design, absolute scoring procedures and a comparative evaluation using the Good–Same–Bad (GSB) metric, with the summarized results provided in Figure~\ref{fig:gsb}. Sec.~\ref{sec:imagination} investigates the performance of the model in a spectrum of core functionalities, such as image/video reference generation, video editing, and the synergistic effects that arise from the composition of multiple capabilities. Sec.~\ref{sec:potential} further examines the extended potential of the model, highlighting its proficiency in more advanced interactive and reasoning-enhanced generative tasks.

\subsection{Human Evaluation} 
\label{sec:human_eval}
\subsubsection{Benchmarks } 
To validate the performance of Kling-Omni compared with other leading video generation and editing models, we constructed the OmniVideo-1.0 Benchmark, which encompasses a comprehensive and representative set of scenarios. We collected a large amount of high-quality multimodal dataset, including images, subjects, and videos as elements. Utilizing this dataset, we designed over 500 cases to comprehensively evaluate the model's capability of referencing, integrating, and editing diverse elements. We meticulously constructed the evaluation set across multiple dimensions, including : \textit{Subject Categories,} which include humans, cartoon characters, animals, clothing, and props; \textit{Application Scenarios,} such as professional video production, e-commerce advertising, and social media content creation; \textit{and Additional Challenges,} involving complex actions, wide-angle perspectives, emotional expressions, cross-style integration, and multi-element fusion.

\begin{figure}[t!]
    \centering
    \includegraphics[width=0.9\textwidth]{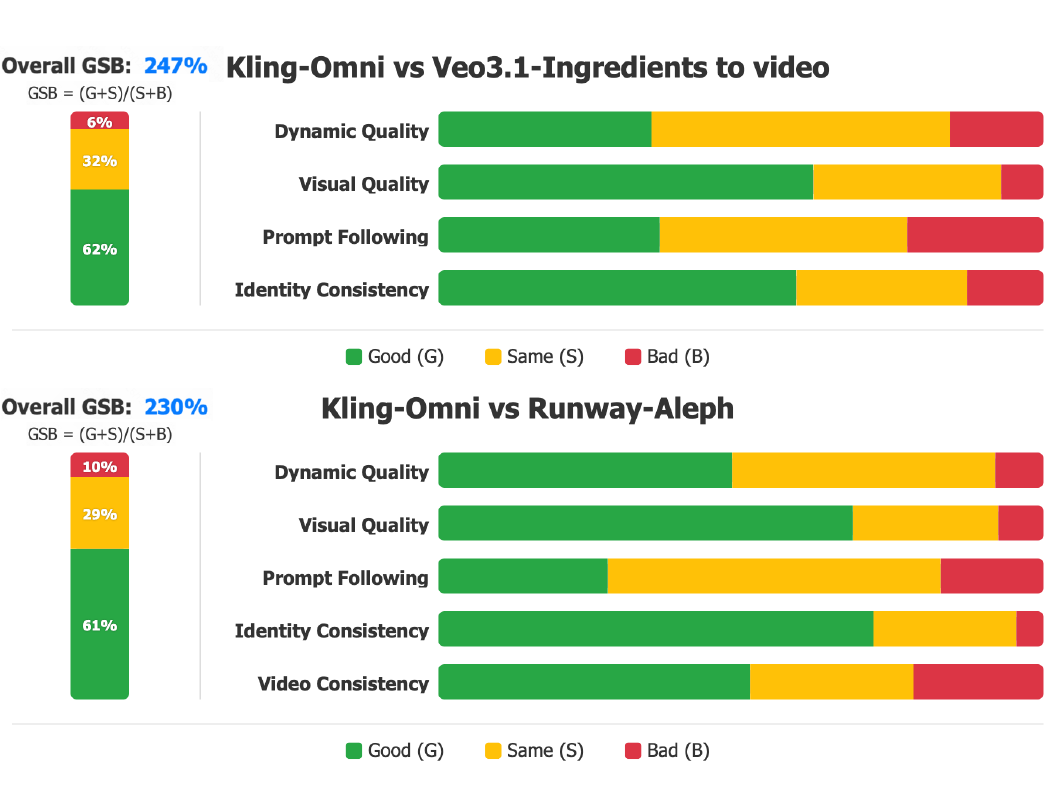}
    \caption{Quantitative comparison of Kling-Omni against SOTA methods on \\ reference-based video generation and video editing tasks. Overall GSB is computed over all evaluation metrics.}
    \label{fig:gsb}
\end{figure}
\subsubsection{Metrics}
We engaged with creators ranging from professional directors to general users. By collecting the requirements from different user groups, we constructed an evaluation system that is comprehensive, structured, and interpretable to evaluate the overall capabilities of model. This system primarily comprises the following core metrics:

\textbf{Dynamic Quality.} This metric assesses the temporal performance of the model, focusing on the continuity between frames, the stability of attributes, and the plausibility of motion relative to physical laws and commonsense dynamics. It evaluates the naturalness of movement, the seamlessness of subject–background integration, and the adequacy of motion amplitude. Additionally, it considers higher-level behaviors, such as multi-character interactions and narrative camera movements, ensuring the generated sequence exhibits stable, realistic, and expressive motion.

\textbf{Prompt Following.}  Reflecting the user's creative intent, this metric measures the model's instruction adherence. It evaluates how accurately the generated video captures and executes the semantic information and specific constraints detailed in the input prompt.

\textbf{Identity Consistency.} This metric evaluates the model's ability to preserve the identity and structural features of reference subjects (e.g., persons, objects, or styles). It assesses stability across variations in camera angles, expressions, complex movements, and lighting conditions throughout the video.

\textbf{Video Consistency}: Specific to video editing tasks, this metric measures the model's faithfulness to unedited regions. It assesses the ability to sustain the identity and structure of key subjects while performing content modifications or style transfers, ensuring visual smoothness and coherence alongside the execution of editing instructions.

\begin{table}[t]
\centering
\renewcommand{\arraystretch}{1.2} 
\caption{Comparison of model capabilities: Kling-Omni vs. SOTA Video Generation and Editing Models.}
\label{tab:capacity_comparison}

\newcolumntype{Y}{>{\centering\arraybackslash}X}

\begin{tabularx}{\textwidth}{llYYY}
\toprule
\textbf{Category} & \textbf{Capability} & \textbf{Kling-Omni (Ours)} & \textbf{Google Veo 3.1} & \textbf{Runway Aleph} \\
\midrule

\multirow{3}{*}{\shortstack[l]{Image/Element Library\\Reference}} 
 & Image Reference & \cmark & \cmark & \xmark \\
 & Element Library Reference & \cmark & \xmark & \xmark \\
 & Image + Element Library Ref. & \cmark & \xmark & \xmark \\
\midrule

\multirow{11}{*}{\shortstack[l]{Instruction\\Editing}} 
 & Addition & \cmark & \cmark & \cmark \\
 & Removal & \cmark & \cmark & \cmark \\
 & Replacement & \cmark & \xmark & \cmark \\
 & Stylization & \cmark & \xmark & \cmark \\
 
 & Attribute Manipulation & \cmark & \xmark & \cmark \\
 & Special Effects & \cmark & \xmark & \cmark \\
 & Video Matting & \cmark & \xmark & \cmark \\
 & Multi-image Editing & \cmark & \xmark & \xmark \\
 & Subject-driven Editing & \cmark & \xmark & \xmark \\
\midrule

\multirow{4}{*}{\shortstack[l]{Video\\Reference}} 
 & Next Shot Generation & \cmark & \xmark & \cmark \\
 & Prev. Shot Generation & \cmark & \xmark & \cmark \\
 & New Camera Angle Generation & \cmark & \xmark & \cmark \\
 & Motion Transfer & \cmark & \xmark & \xmark \\
 & Camera Motion Transfer & \cmark & \xmark & \xmark \\
\midrule

\multirow{2}{*}{\shortstack[l]{Frame-cond.\\Generation}} 
 & First Frame to Video & \cmark & \cmark & \cmark \\
 & First \& Last Frame & \cmark & \cmark & \cmark \\
\midrule

\multicolumn{2}{l}{Text-to-Video} & \cmark & \cmark & \cmark \\
\midrule
\multicolumn{2}{l}{Compositional Generation} & \cmark & \xmark & \cmark \\
\midrule
\multicolumn{2}{l}{Visuao Prompt Understanding} & \cmark & \cmark & \xmark \\
\midrule
\multicolumn{2}{l}{Reasoning-enhanced Generation} & \cmark & \cmark & \xmark \\
\bottomrule
\end{tabularx}
\end{table}

\subsubsection{Evaluation Results}
We conducted a double-blind human evaluation based on the OmniVideo-Benchmark 1.0, inviting domain experts and professional annotators to compare Kling-Omni against industry-leading models. Evaluators performed side-by-side qualitative assessments based on the defined dimensions, classifying the relative performance into three categories:

\textbf{G (Good):} The performance of Kling-Omni is significantly superior to the competing model.

\textbf{S (Same):} The performance of Kling-Omni is comparable to the competing model.

\textbf{B (Bad):} The performance of Kling-Omni is significantly inferior to the competing model.

The aggregated GSB metric distributions for Image-Reference and Video-Editing tasks are presented in Fig. ~\ref{fig:gsb}. We compared Kling-Omni against Veo 3.1~\cite{veo3} for image-referencing tasks and Runway-Aleph~\cite{runway_aleph} for video editing tasks. As illustrated, Kling-Omni demonstrates varying degrees of superiority over its competitors across all evaluated dimensions, validating its robustness and reliability in complex generation and editing scenarios.

\subsection{Unleash Imagination via Kling-Omni}
\label{sec:imagination}
This section demonstrates the capabilities of Kling-Omni. Table~\ref{tab:capacity_comparison} lists the representative features, including but not limited to reference-based generation, instruction-driven editing, video reference, frame-condition generation, compositional generation, visual prompt understanding, intelligent reasoning via generation, etc. Qualitative analysis for representative features  are provided below.
\subsubsection{Multi-Modal and Multi-Dimensional Precise Referencing}

Kling-Omni enables fine-grained and reliable control through multi-modal and multi-dimensional referencing, as shown in Table~\ref{tab:capacity_comparison}. The model supports flexible conditioning based on diverse input forms—images, videos, and text—and allows users to specify reference information across multiple dimensions, including but not limited to identity, status, style, shot composition, and actions. Unlike single-image-per-subject referencing approaches, Kling-Omni incorporates a subject library mechanism, where multiple images of the same subject (e.g., the same person with different viewpoints, poses, expressions, or lighting conditions) can be jointly provided. This design improves the ability of the model to establish a consistent and robust subject representation, allowing more stable identity preservation.

\begin{figure}[t!]
    \centering
    \includegraphics[width=0.9\textwidth]{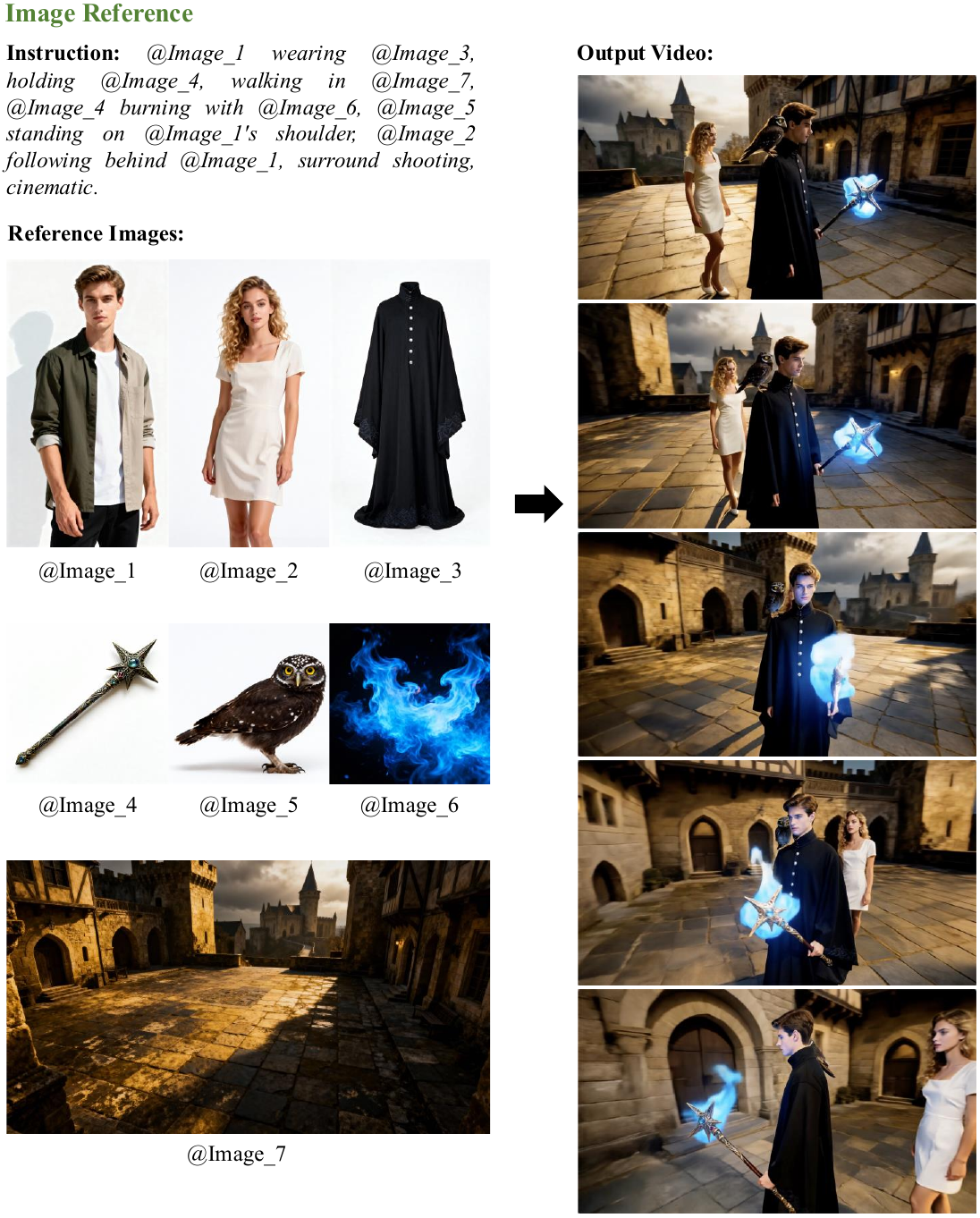}
    \caption{Examples of image-reference-based video generation.}
    \label{fig:video_multiref_image}
\end{figure}

\begin{figure}[t!]
    \centering
    \includegraphics[width=0.9\textwidth]{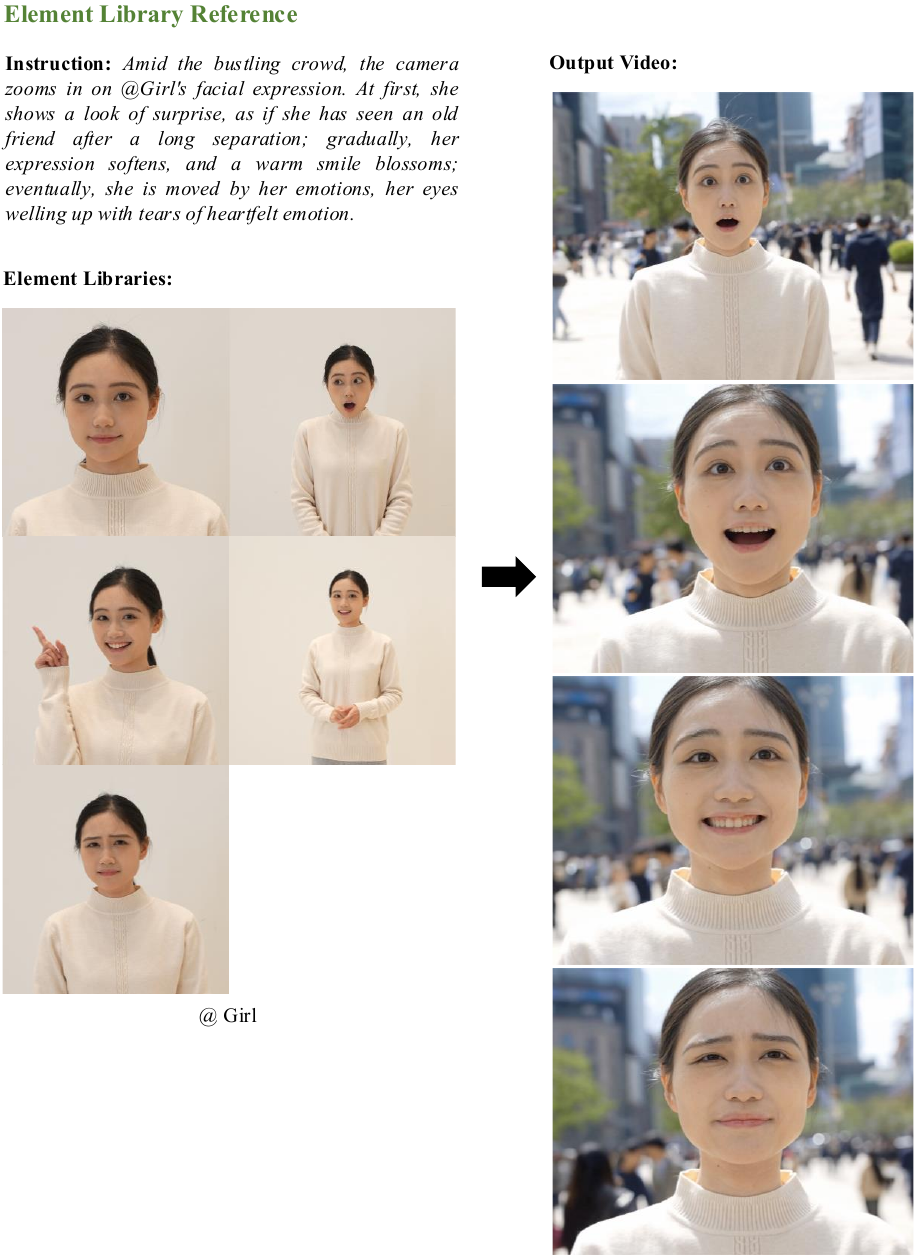}
    \caption{Examples of element library reference. Kling-Omni supports multi-expression references for the same subject.}
    \label{fig:video_multiref_object_emotion}
\end{figure}

\begin{figure}[t!]
    \centering
    \includegraphics[width=0.9\textwidth]{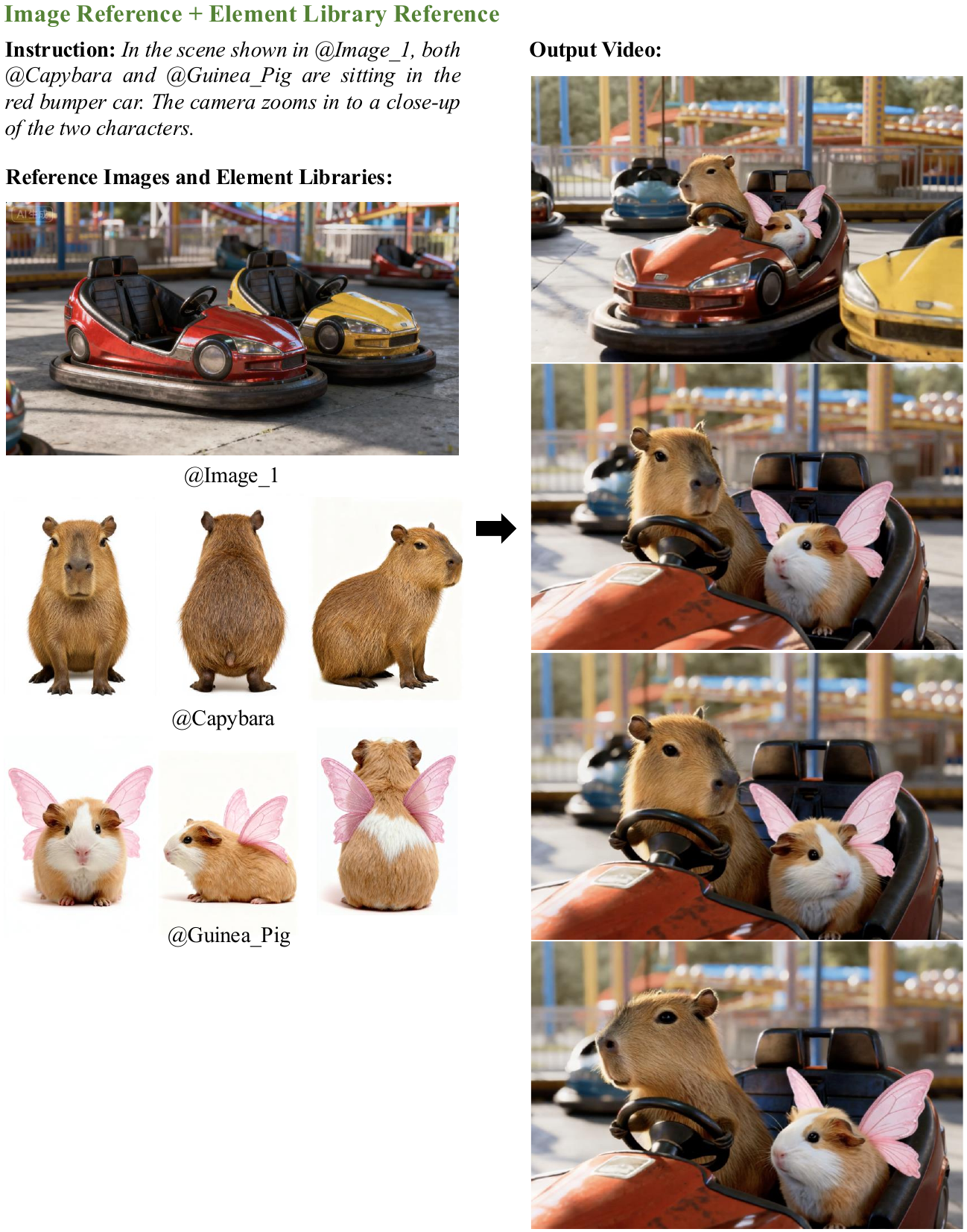}
    \caption{Examples of image reference together with element library reference.}
    \label{fig:video_multiref_object_view}
\end{figure}

\begin{figure}[t!]
    \centering
    \includegraphics[width=0.9\textwidth]{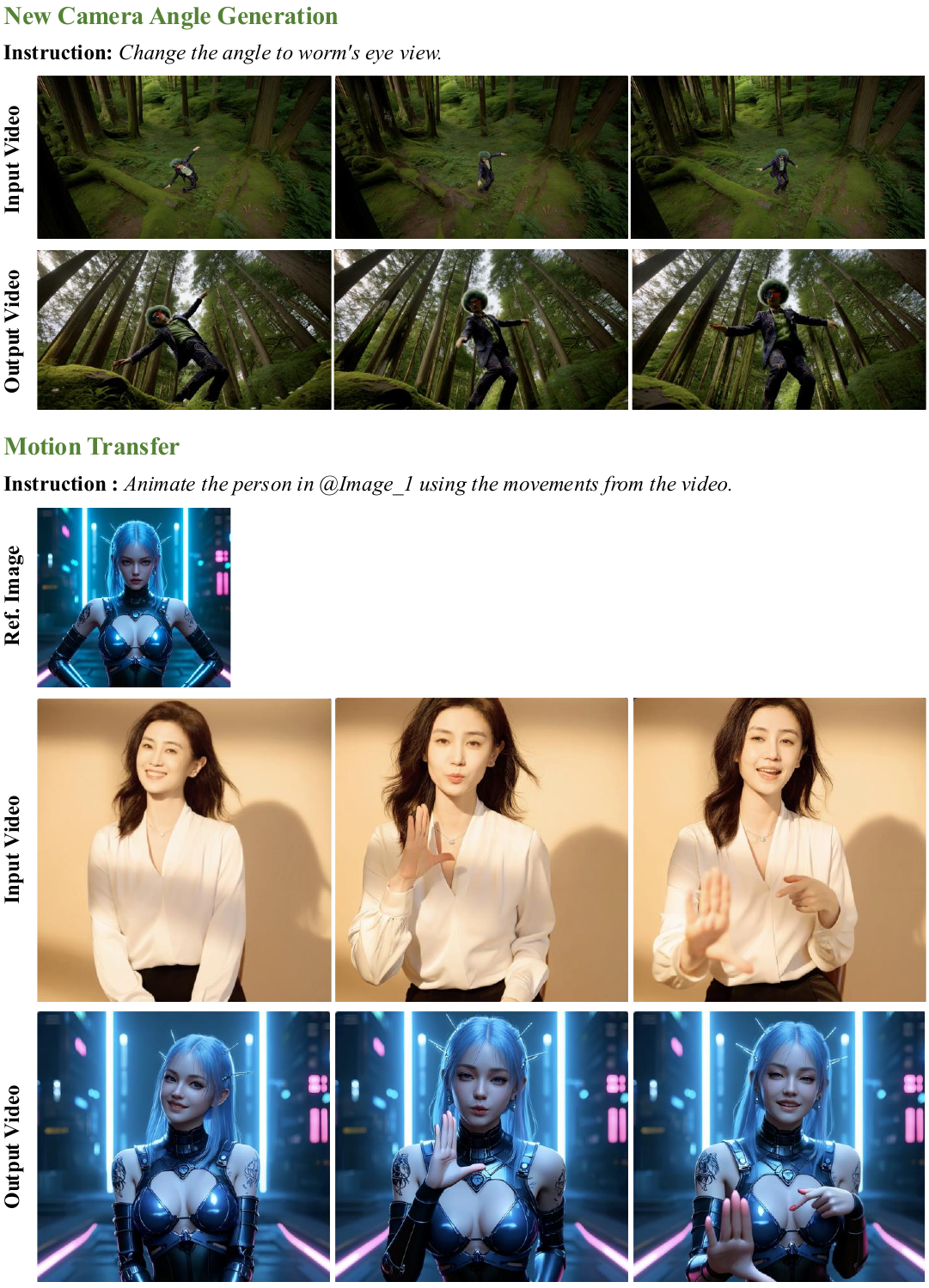}
    \caption{Examples of new camera angle generation and motion transfer in video reference.}
    \label{fig:video_reference_part1}
\end{figure}

\begin{figure}[t!]
    \centering
    \includegraphics[width=0.9\textwidth]{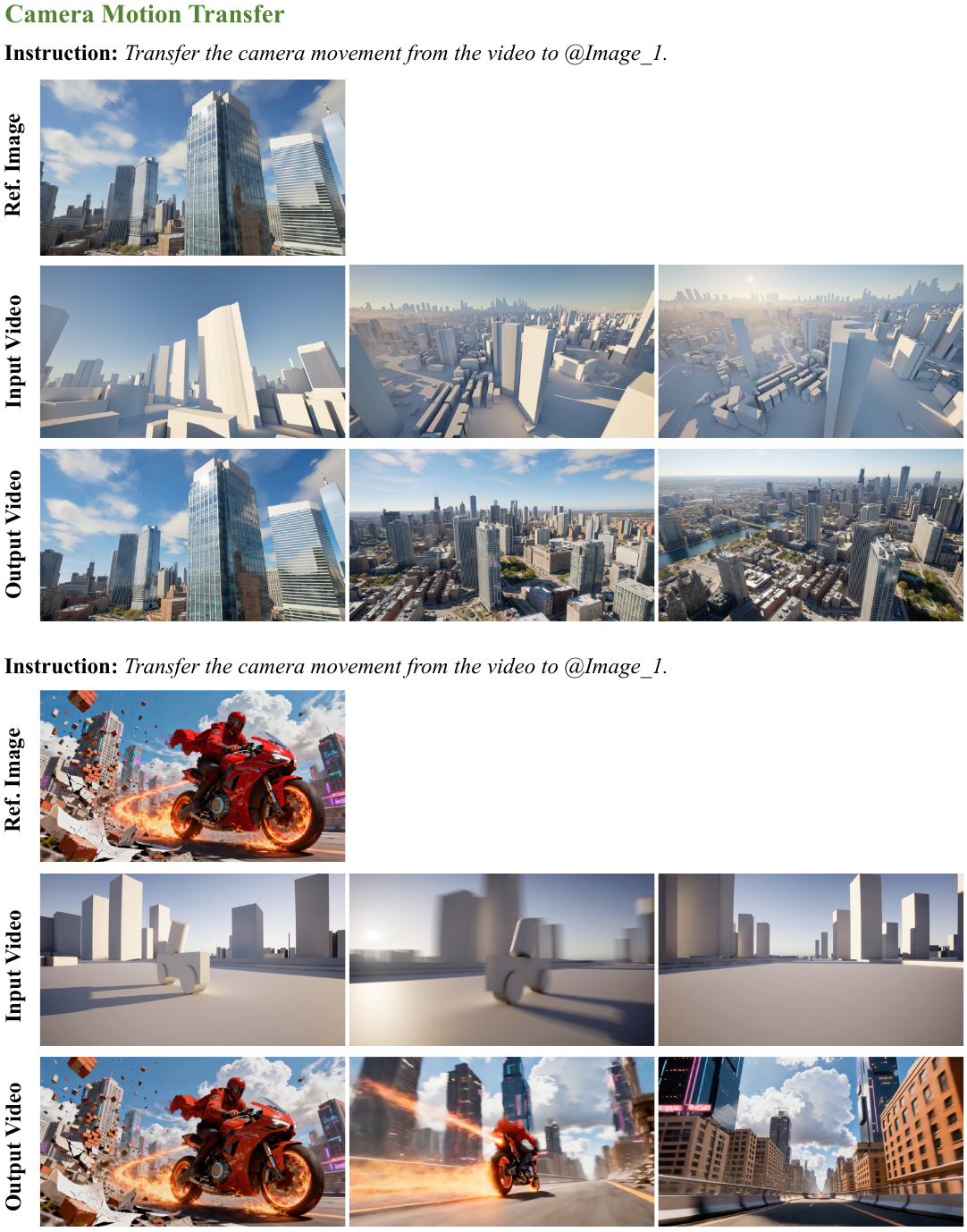}
    \caption{Examples of camera motion transfer in video reference.}
    \label{fig:video_reference_part2}
\end{figure}

\begin{figure}[t!]
    \centering
    \includegraphics[width=0.9\textwidth]{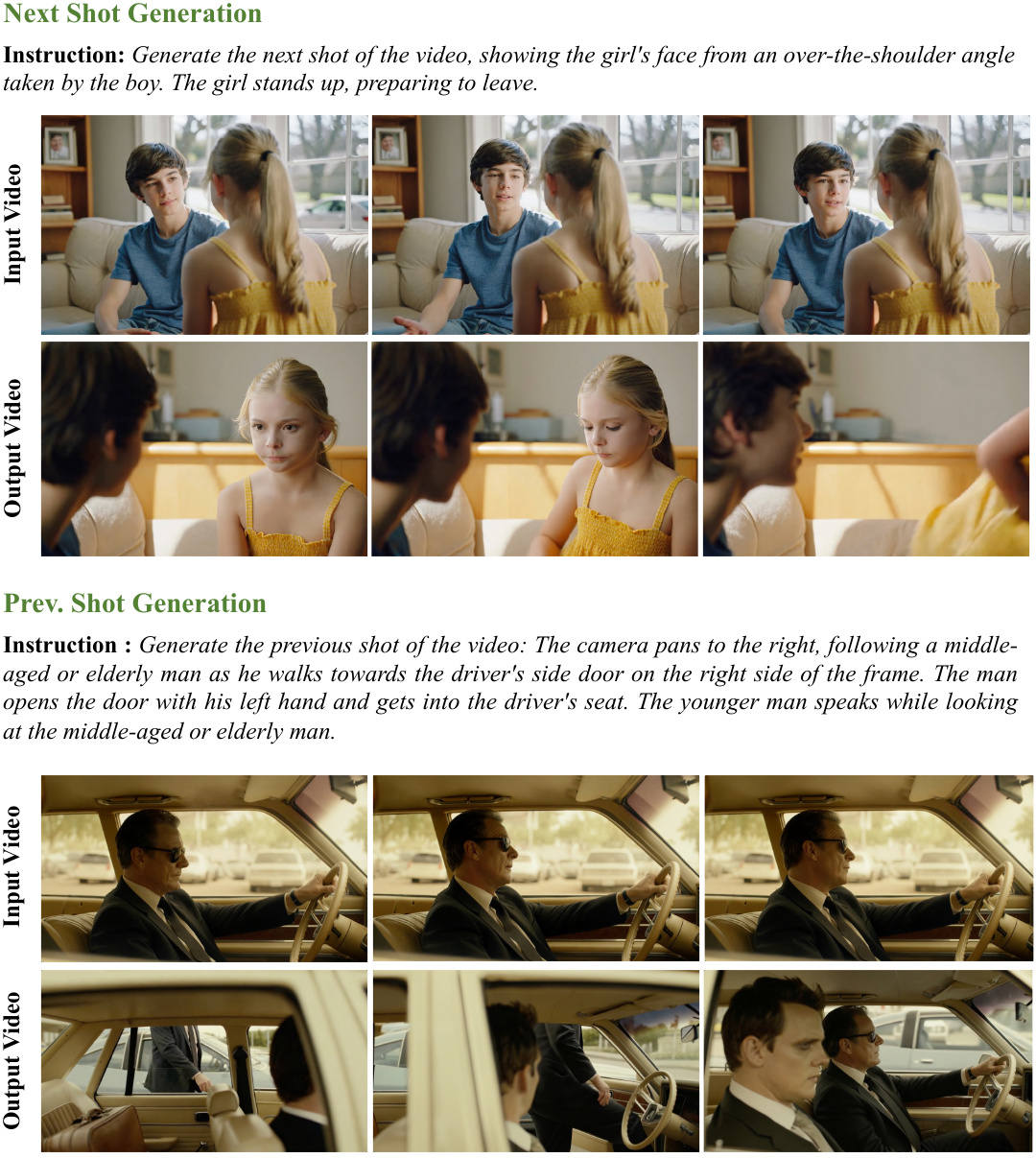}
    \caption{Examples of next shot generation and previous shot generation in video reference.}
    \label{fig:video_reference_part3}
\end{figure}

\begin{figure}[t!]
    \centering
    \includegraphics[width=0.9\textwidth]{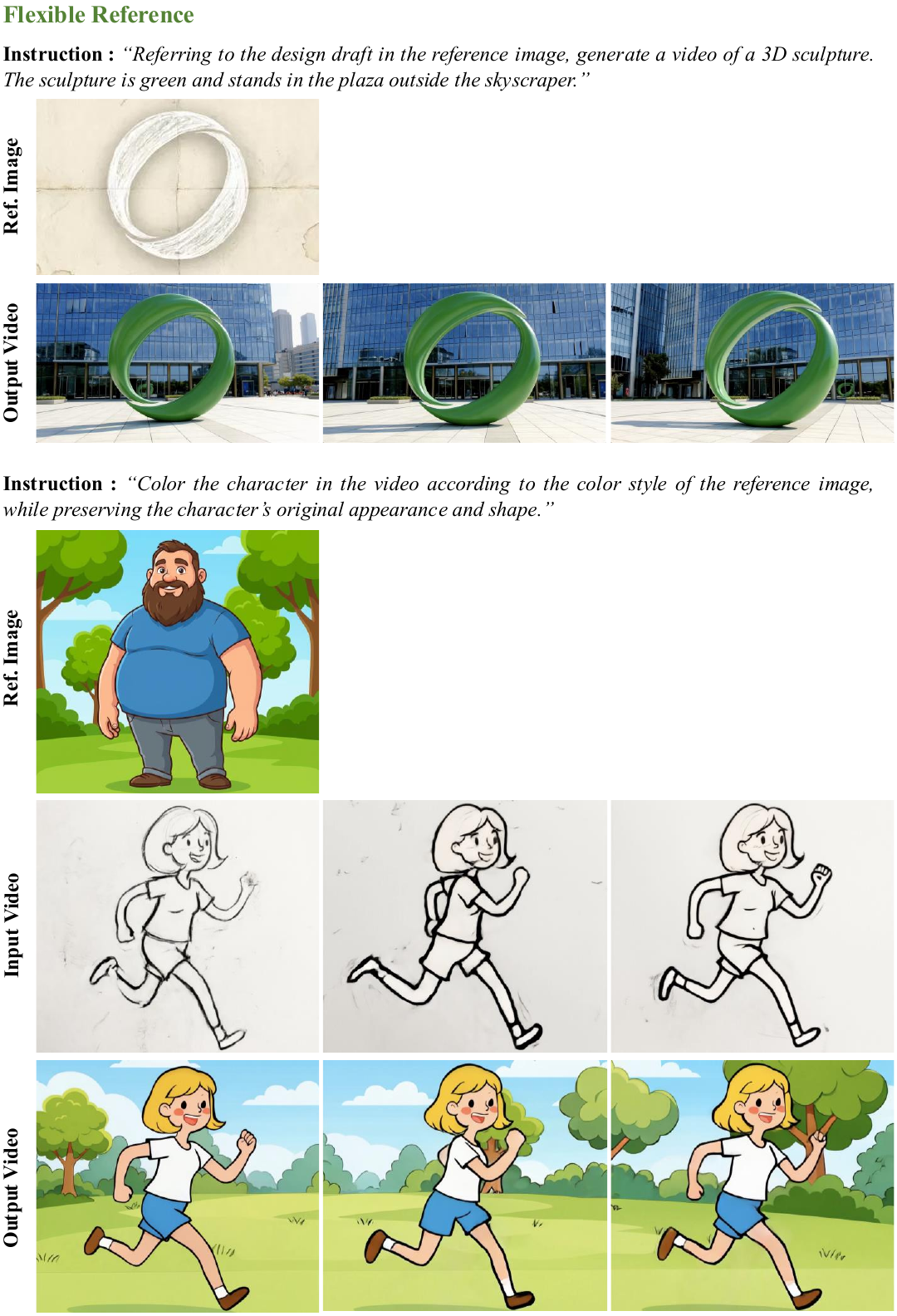}
    \caption{Examples of flexible image and video reference, e.g, sketch reference. The top example shows video generation controlled by the sketch drawing in the reference image, while the bottom example illustrates video stylization that integrates color references into the sequential sketch reference of a video.}
    \label{fig:flexible_reference}
\end{figure}

By integrating these multi-source and multi-dimensional cues, Kling-Omni achieves precise alignment with user intent while maintaining visual and semantic coherence in complex generation scenarios, including image/element library reference, new camera angle generation, motion transfer, camera motion transfer, next-shot generation, previous-shot generation and flexible referencing dimension like sketch, as shown in Fig.~\ref{fig:video_multiref_image} to Fig.~\ref{fig:flexible_reference}. 
This flexible referencing paradigm also leaves room for users to explore richer combinations of reference dimensions beyond the predefined set, further exploring the potential of Kling-Omni.

\subsubsection{Temporal Narrative}

\begin{figure}[t!]
    \centering
    \includegraphics[width=0.9\textwidth]{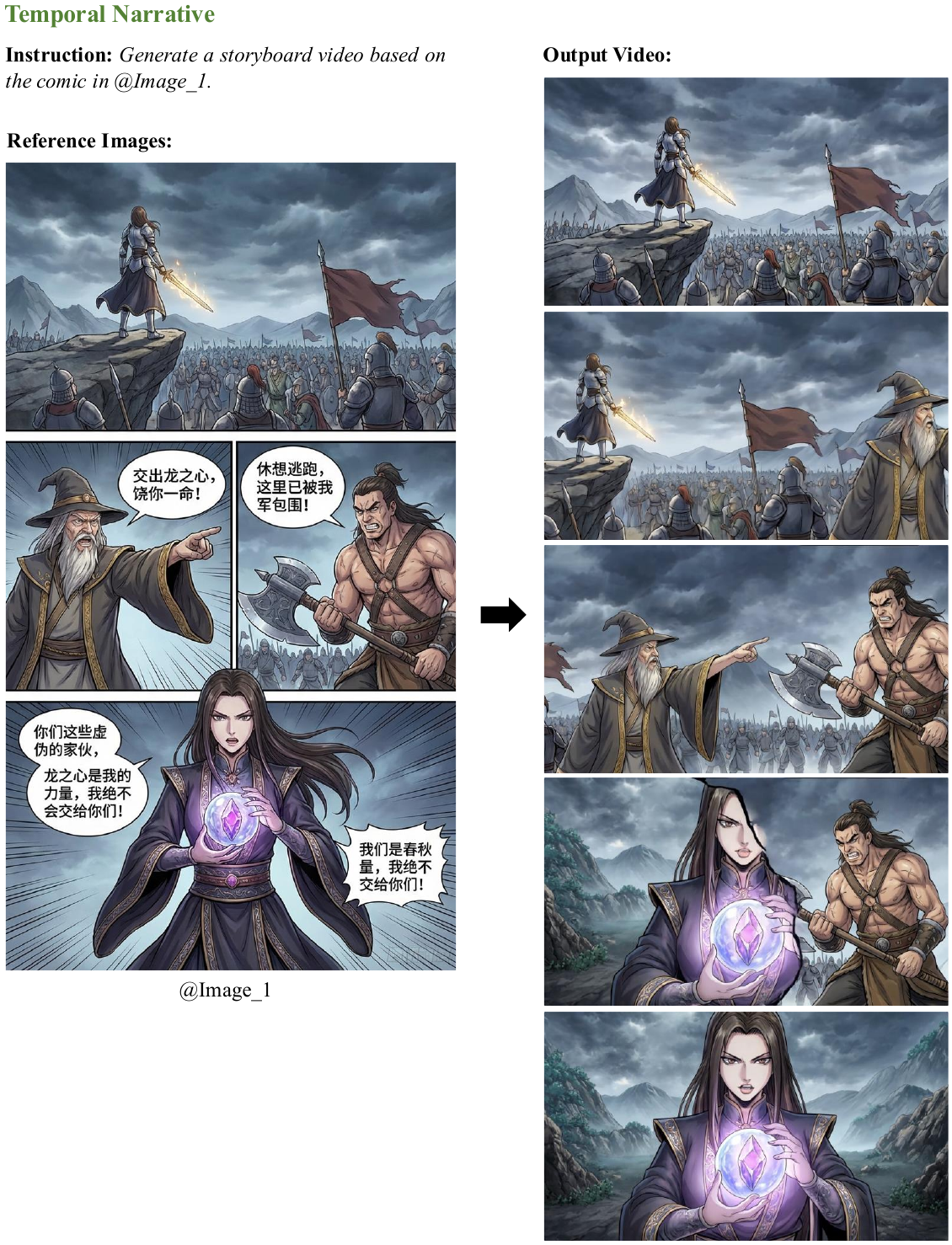}
    \caption{Examples of temporal narrative in image reference. The input is a multi-grid image.}
    \label{fig:video_multiref_image_temporal_case1}
\end{figure}

\begin{figure}[t!]
    \centering
    \includegraphics[width=0.9\textwidth]{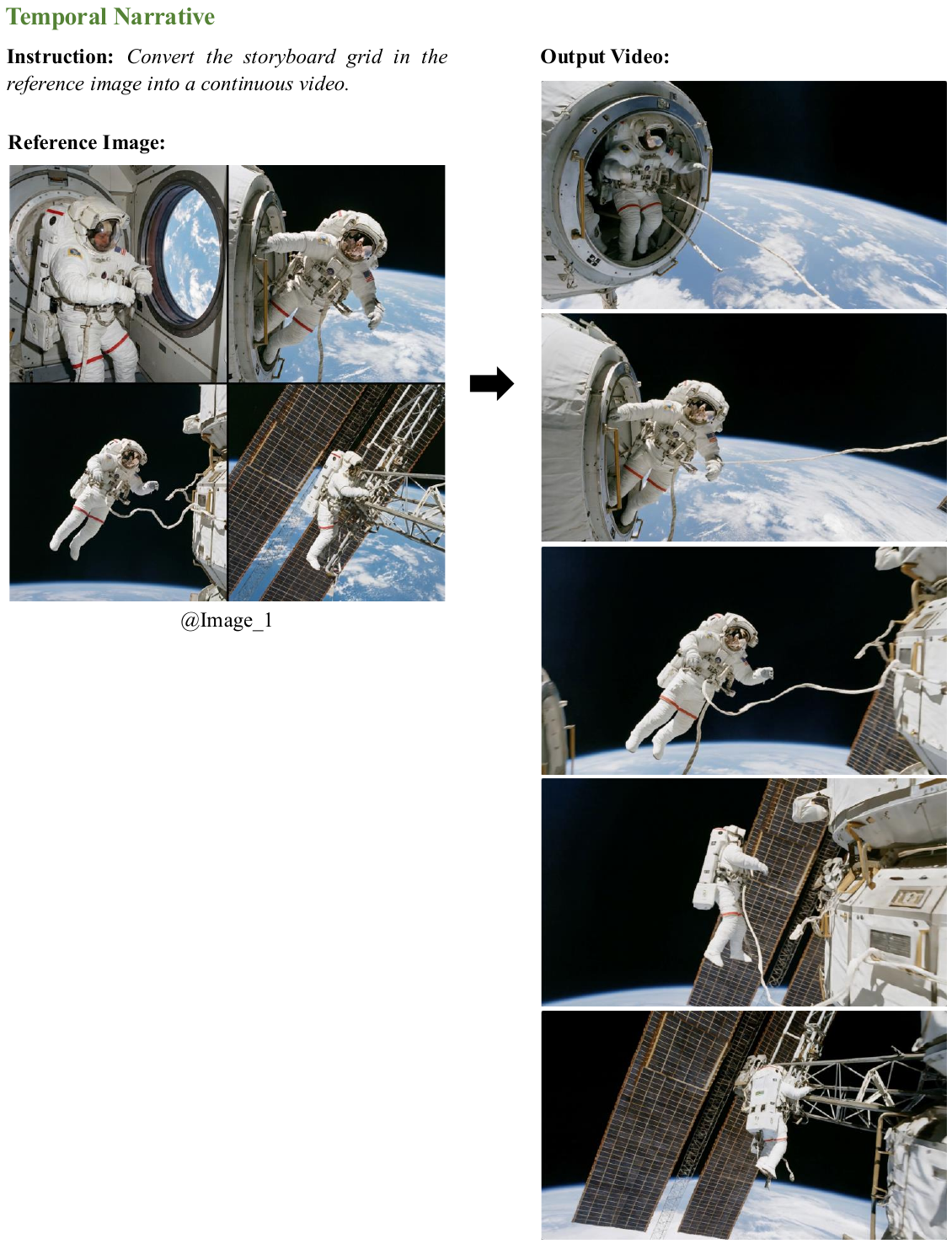}
    \caption{Examples of temporal narrative in image reference. The input is a multi-grid image.}
    \label{fig:video_multiref_image_temporal_case2}
\end{figure}

This feature enables the model to interpret a group of related images—whether they depict a continuous single shot or a complex multi-shot sequence—and generate a comprehensive video presentation, as shown in Fig.~\ref{fig:video_multiref_image_temporal_case1} and Fig.~\ref{fig:video_multiref_image_temporal_case2}. By intelligently bridging the visual gaps between frames, the model constructs a cohesive, chronologically flowing narrative that transforms a static storyboard into a dynamic video experience.

\subsubsection{High-Degree-of-Freedom Interactive Editing}

\begin{figure}[t!]
    \centering
    \includegraphics[width=0.9\textwidth]{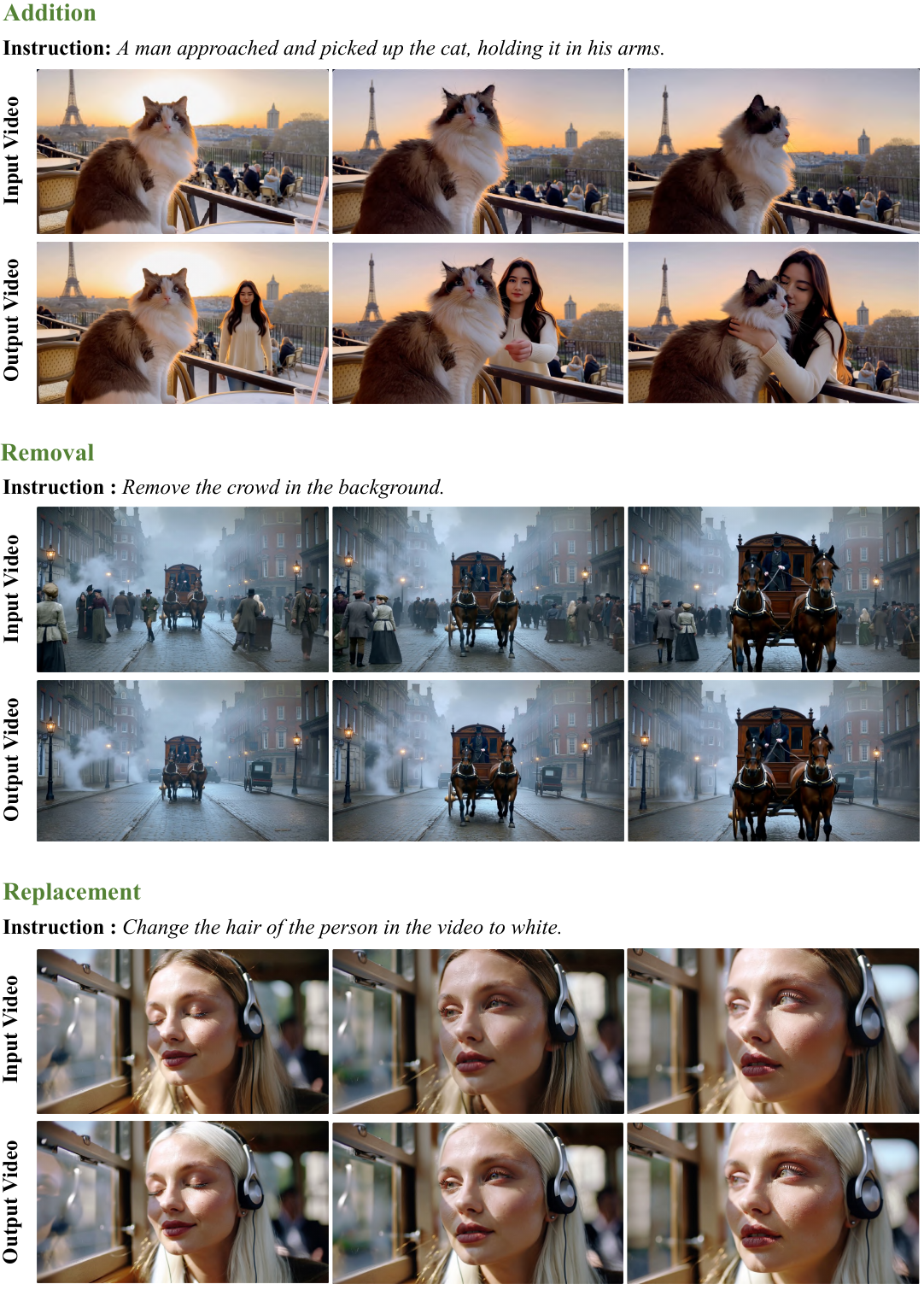}
    \caption{Examples of addition, removal, and replacement in video editing.}
    \label{fig:video_editing_noref}
\end{figure}

\begin{figure}[t!]
    \centering
    \includegraphics[width=0.9\textwidth]{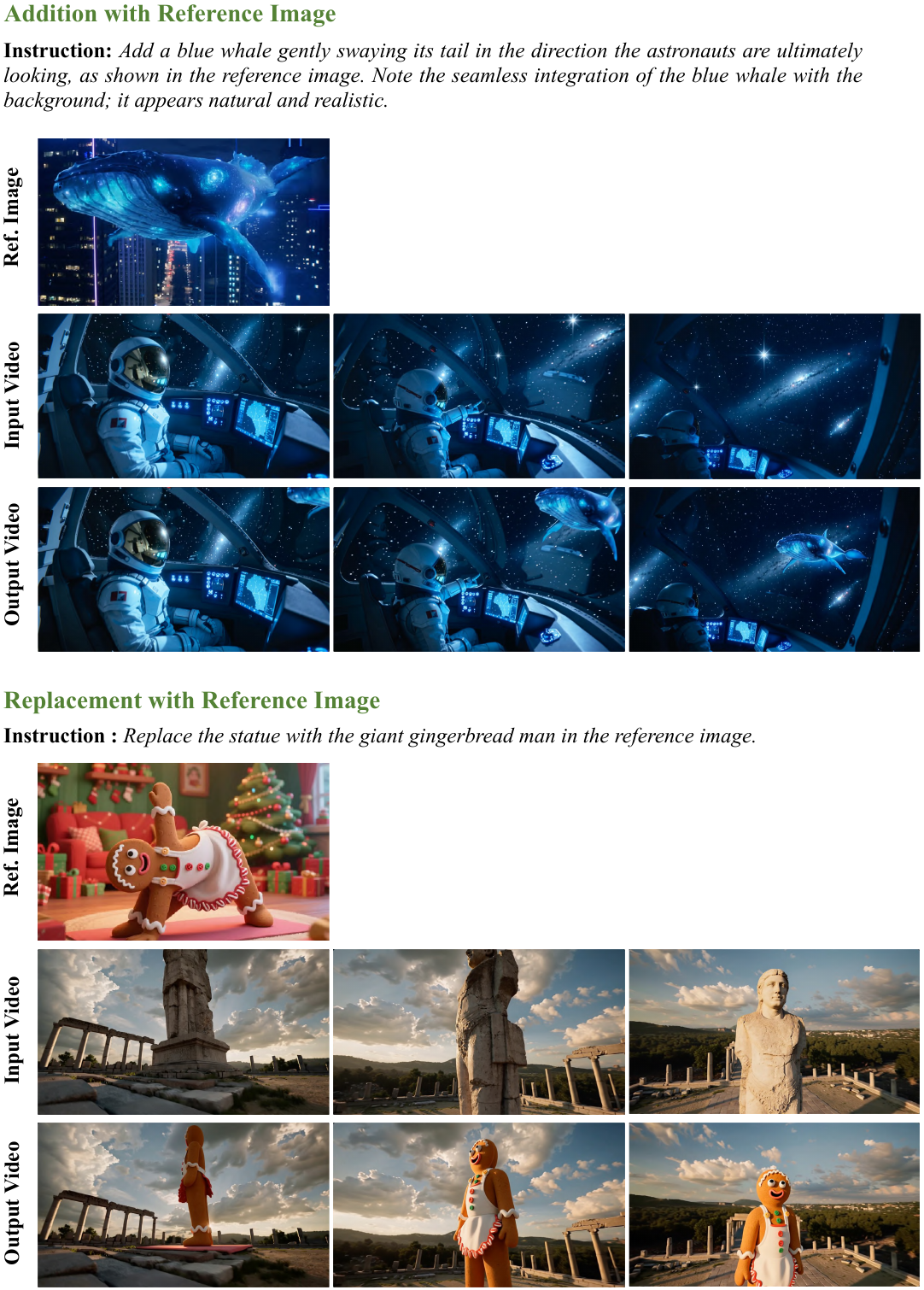}
    \caption{Examples of reference-image–guided addition and replacement in video editing.}
    \label{fig:video_editing_wref}
\end{figure}

\begin{figure}[t!]
    \centering
    \includegraphics[width=0.9\textwidth]{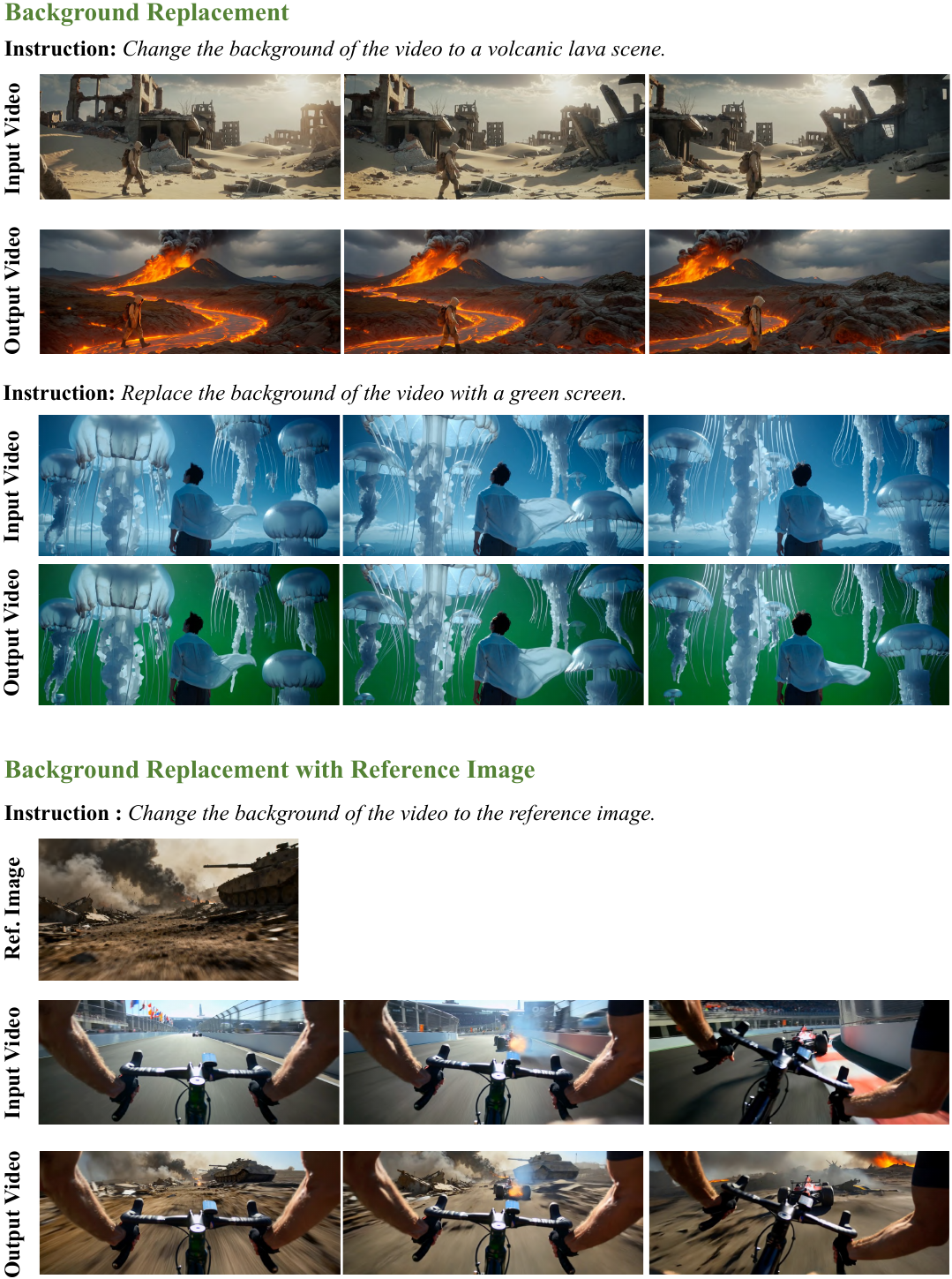}
    \caption{Examples of background replacement in video editing.}
    \label{fig:video_editing_background}
\end{figure}

\begin{figure}[t!]
    \centering
    \includegraphics[width=0.9\textwidth]{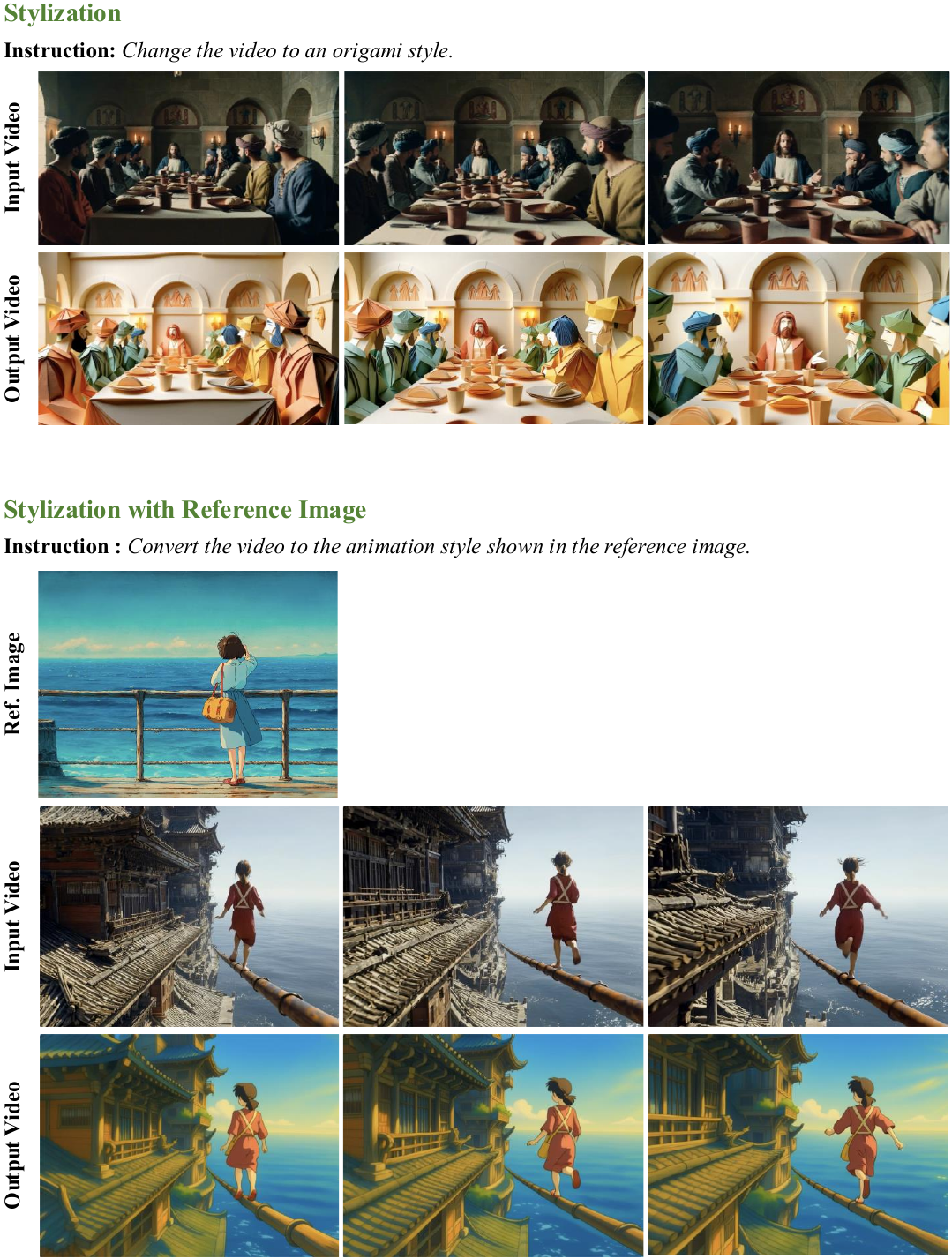}
    \caption{Examples of video stylization in video editing.}
    \label{fig:video_editing_style}
\end{figure}

\begin{figure}[t!]
    \centering
    \includegraphics[width=0.9\textwidth]{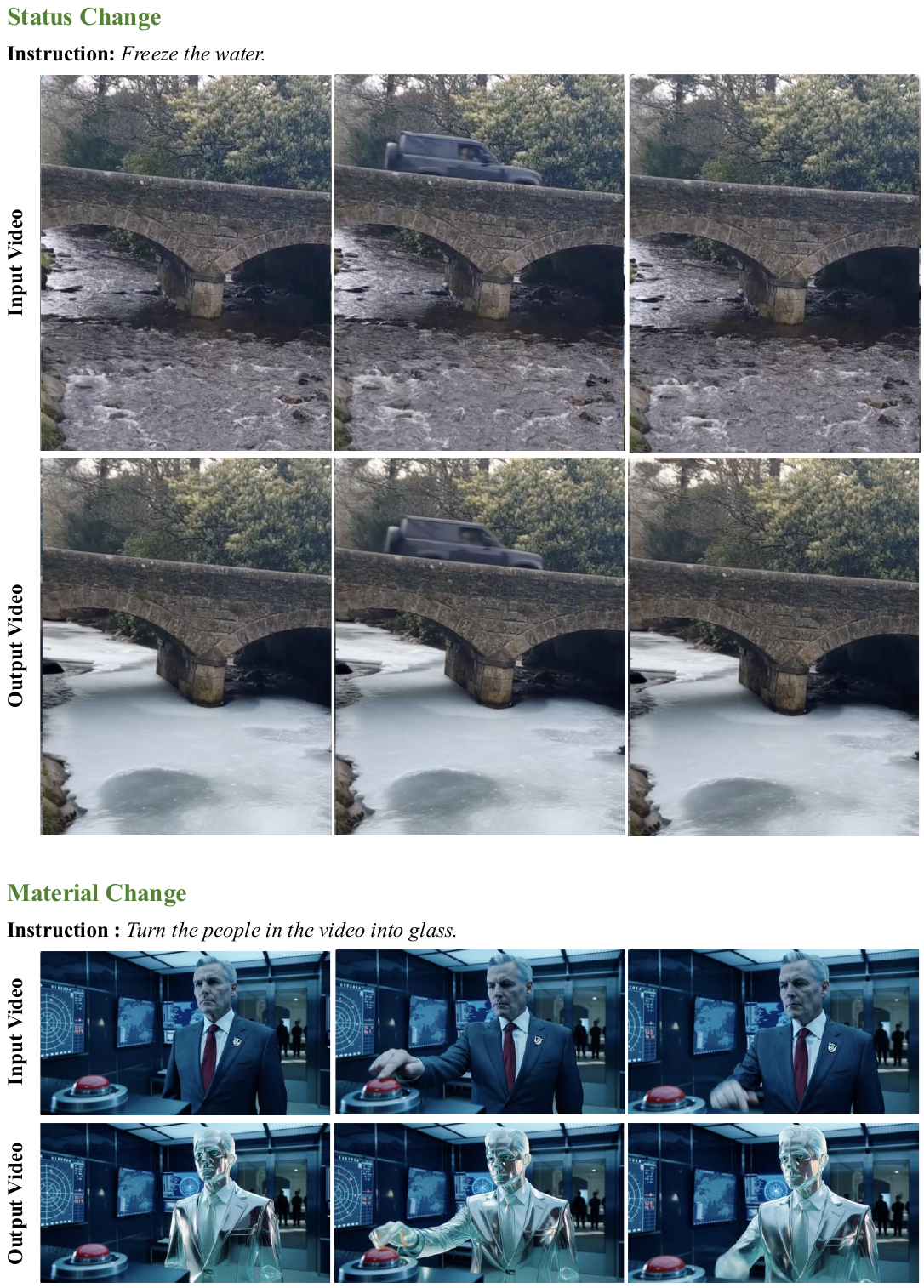}
    \caption{Examples of attribute manipulation in video editing.}
    \label{fig:video_editing_flexible_part1}
\end{figure}

\begin{figure}[t!]
    \centering
    \includegraphics[width=0.9\textwidth]{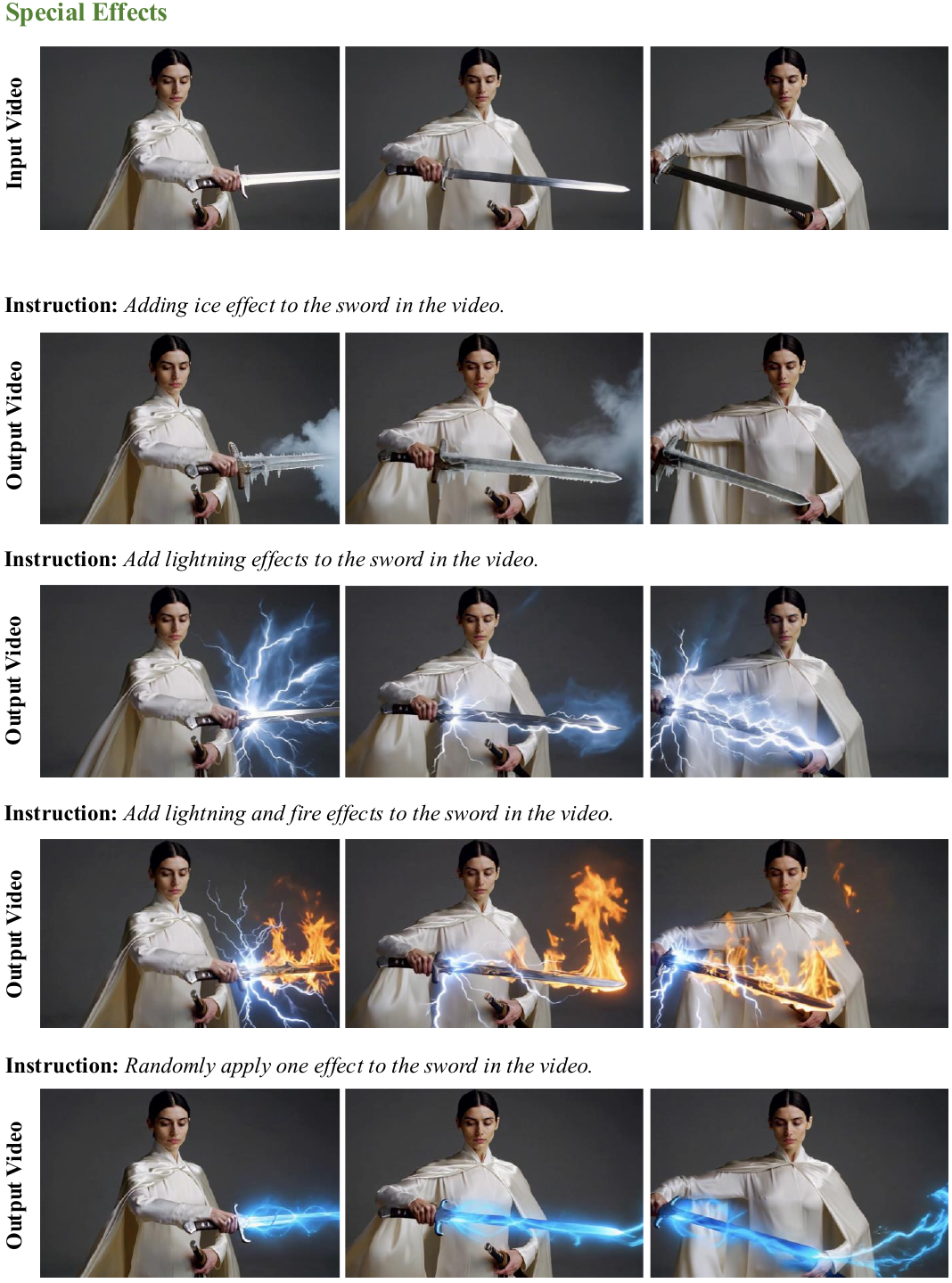}
    \caption{Examples of special effects in video editing.}
    \label{fig:video_editing_vfx}
\end{figure}

\begin{figure}[t!]
    \centering
    \includegraphics[width=0.9\textwidth]{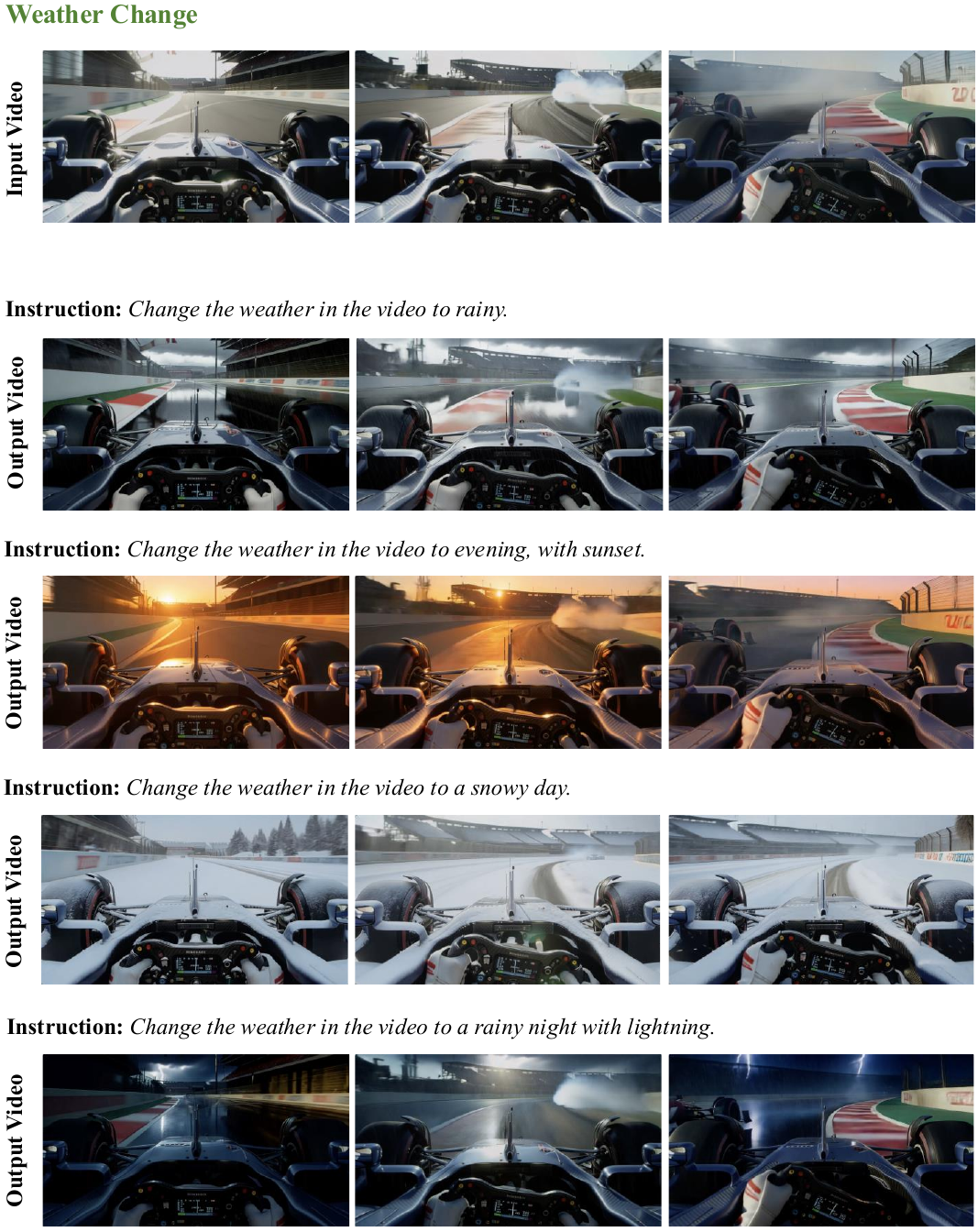}
    \caption{Examples of weather change in video editing.}
    \label{fig:video_editing_weather}
\end{figure}

In addition to conventional edit operations such as addition, removal, and replacement of content, Kling-Omni enables unconstrained interactive manipulation that is free from temporal and spatial limitations, allowing users to control video content along arbitrary dimensions—including elements, styles, scenes, and shots, as shown in Fig.~\ref{fig:video_editing_noref} to Fig.~\ref{fig:video_editing_weather}.

\subsubsection{Flexible Task Combination}
\begin{figure}[t!]
    \centering
    \includegraphics[width=0.9\textwidth]{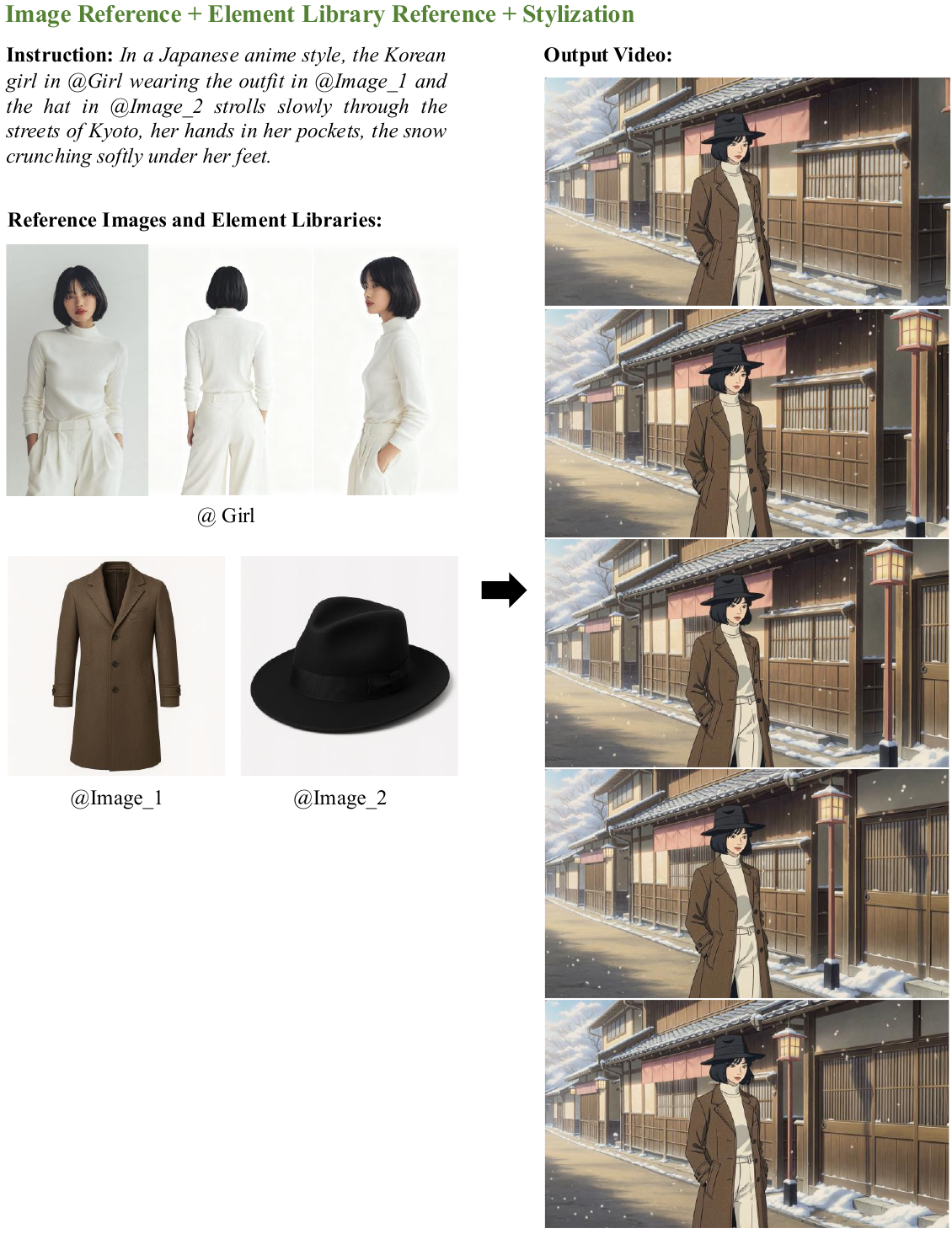}
    \caption{Example of task composition: Kling-Omni combines the element library of a girl, reference images, and an video stylization prompt to generate a consistent stylized video.}
    \label{fig:video_multiref_image_composition}
\end{figure}

\begin{figure}[t!]
    \centering
    \includegraphics[width=0.9\textwidth]{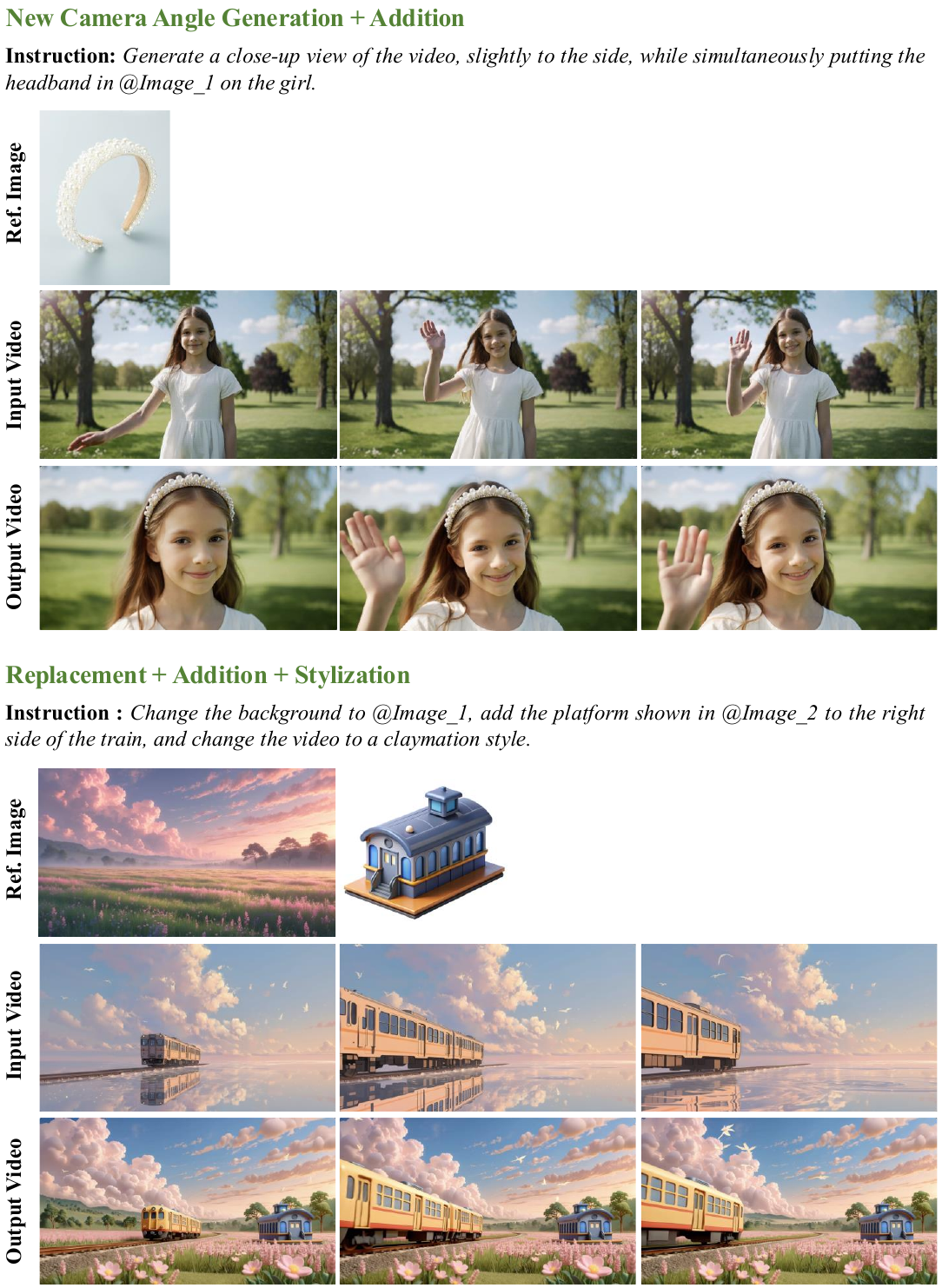}
    \caption{Two examples of task composition: (top) generating a new camera angle while adding a referenced headband; (bottom) replacing the background, adding a train platform element, and converting the video to a claymation style.}
    \label{fig:video_editing_composition}
\end{figure}

As shown in Fig.~\ref{fig:video_multiref_image_composition} and Fig.~\ref{fig:video_editing_composition}, the model has the ability to handle combined complex instructions within a single generation process, without requiring sequential task execution or manual decomposition. This unified approach not only simplifies the workflow but also avoids the accumulation of errors that typically occur in sequential editing, ensuring more consistent and accurate results while improving overall generation efficiency.

\subsection[Broader Potential of Kling-Omni]
{Broader Potential of Kling-Omni\footnote{Features described in this section are not yet supported in the online version.}}
\label{sec:potential}
\subsubsection{Controllable Generation via Visual Signals}

\begin{figure}[t!]
    \centering
    \includegraphics[width=0.9\textwidth]{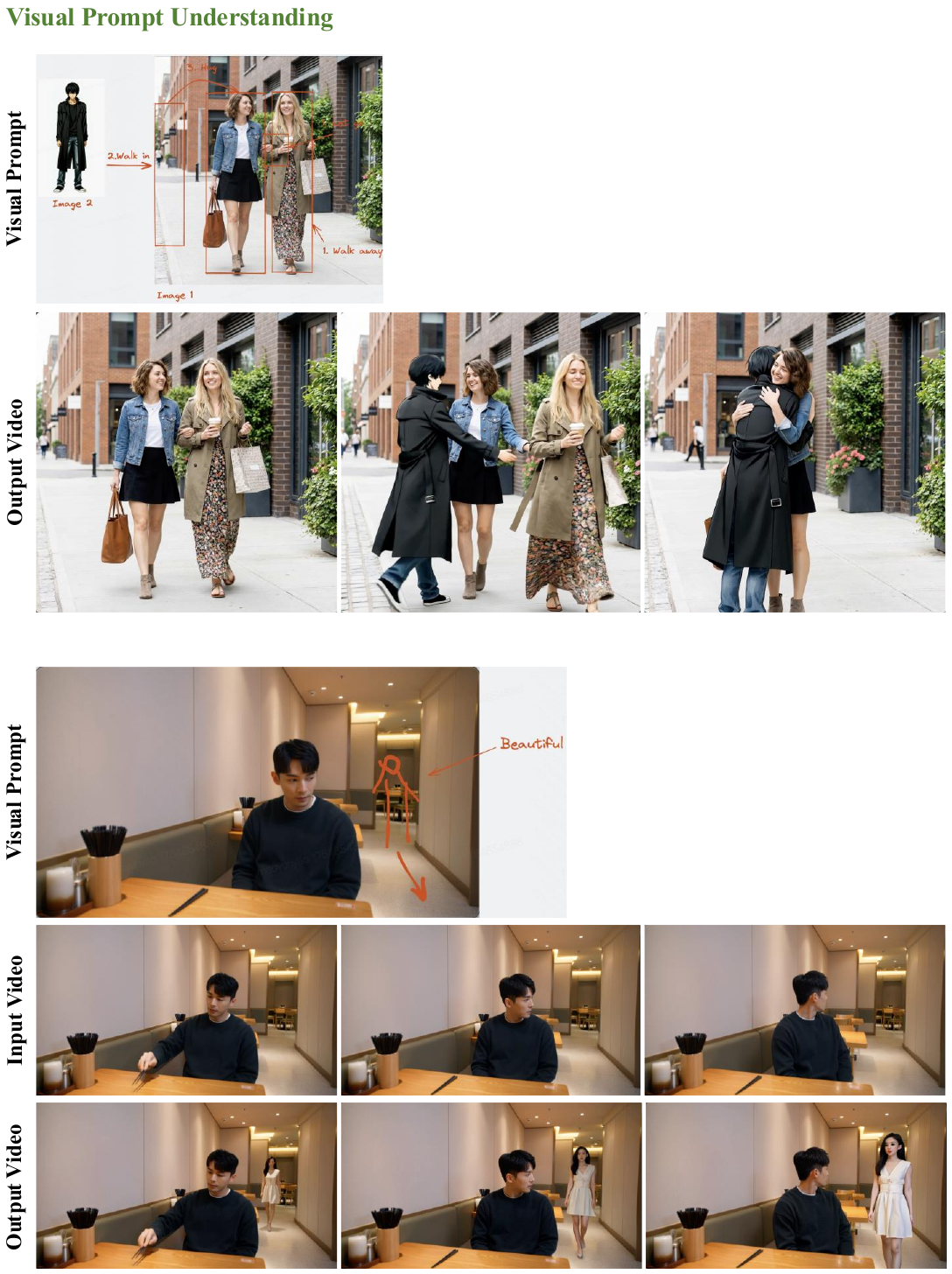}
    \caption{Examples of visual-signal-guided video generation, which supports intelligent interpretation of user intent from images containing visual signals.}
    \label{fig:visual_signal}
\end{figure}

Moving beyond traditional text-based prompting, we conduct an experimental investigation of video generation driven by visual signals. We take advantage of a powerful vision–language reasoning model to explore new possibilities in customized video synthesis. We adopt a workflow in which users express their intent through visual annotations—such as drawing arrows to indicate character trajectories or using bounding boxes to specify interactions. As shown in Fig.~\ref{fig:visual_signal}, by interpreting these visual cues, the model translates abstract user concepts into concrete generation constraints. This example demonstrates the promising potential of advanced vision–language systems to achieve fine-grained control over character identity and scene dynamics.

\subsubsection{Reasoning-enhanced Generation}

\begin{figure}[t!]
    \centering
    \includegraphics[width=0.9\textwidth]{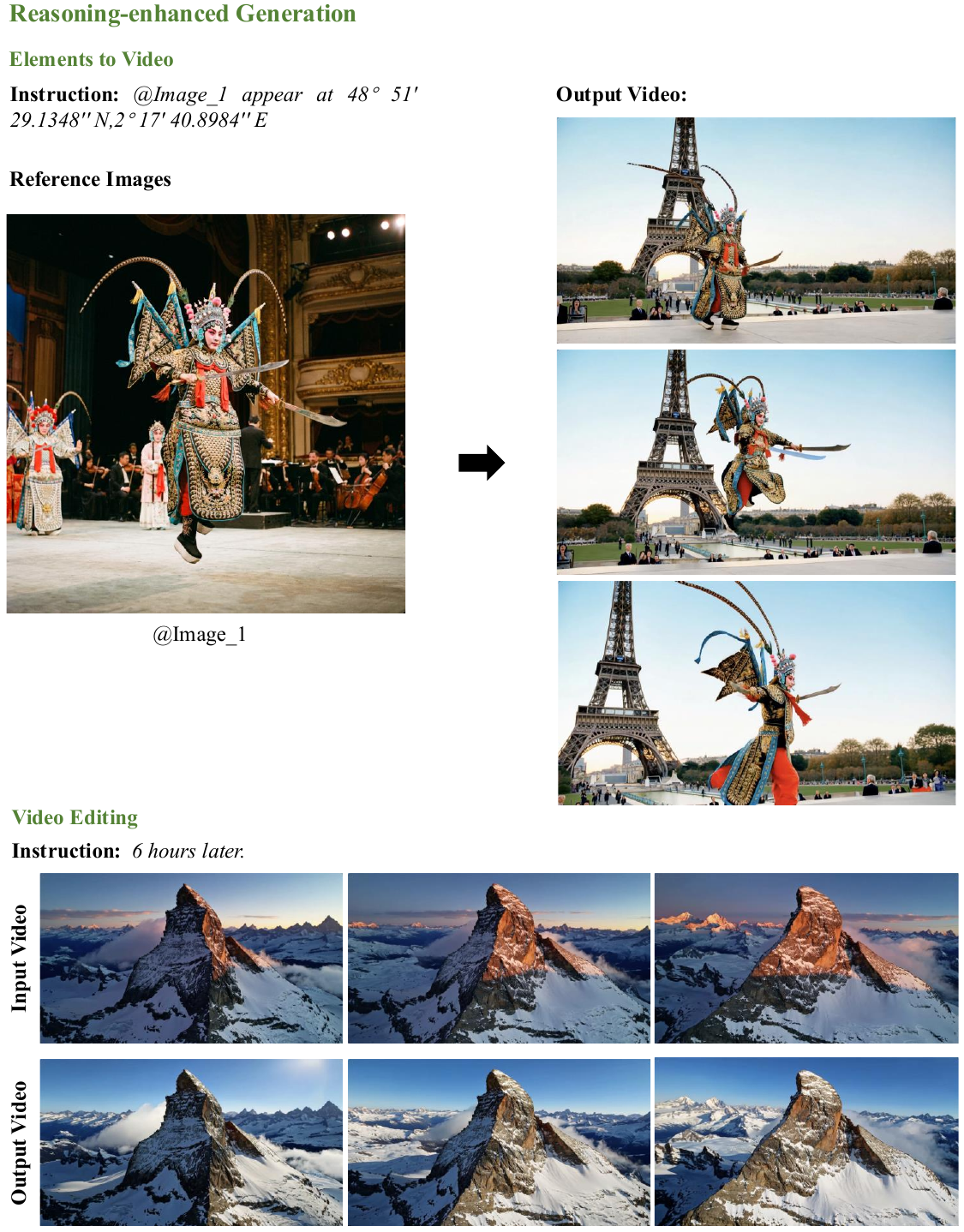}
    \caption{Examples of reasoning-enhanced generation leveraging world knowledge. The top one demonstrates geospatial reasoning by synthesizing a subject into the specific location defined by GPS coordinates (the Eiffel Tower). The bottom one showcases temporal reasoning, where the model accurately adjusts environmental lighting and shadows on a mountain landscape based on the instruction "6 hours later."}
    \label{fig:reasoning_generation_multiref}
\end{figure}

We conduct an exploratory study on intelligent reasoning–enhanced generation, integrating a more powerful vision–language reasoning engine to bridge the gap between abstract user prompts and concrete visual execution. As shown in Fig.~\ref{fig:reasoning_generation_multiref}, the system leverages world knowledge, such as interpreting GPS coordinates or inferring temporal dynamics, to ground user instructions in real-world context. For example, it can decode raw geographic coordinates to retrieve associated landmark knowledge (e.g., the Eiffel Tower), enabling context-aware scene synthesis.

\begin{figure}[t!]
    \centering
    \includegraphics[width=0.9\textwidth]{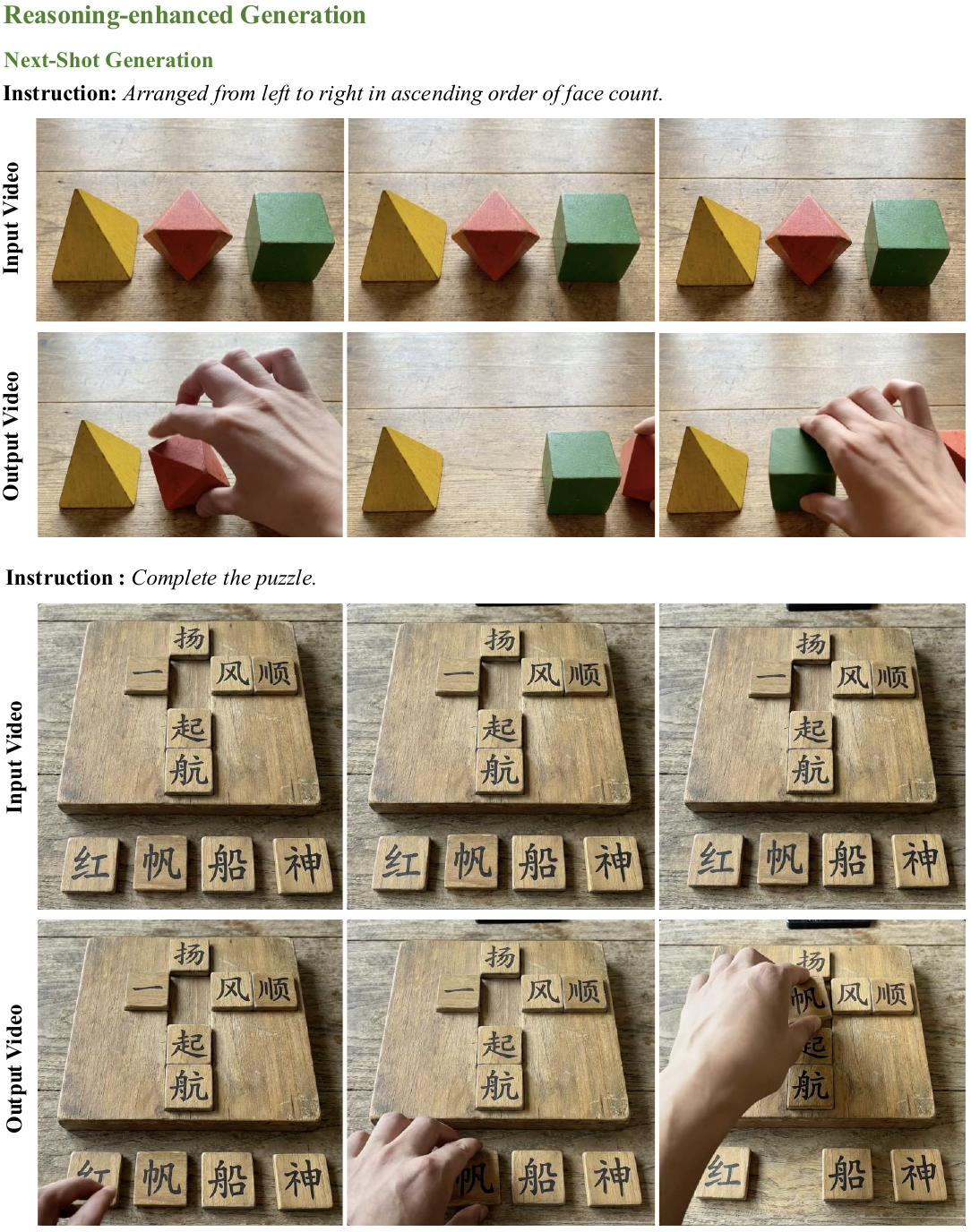}
    \caption{Examples of reasoning-enhanced generation for logical tasks. The top one demonstrates sorting geometric shapes (tetrahedron, cube, octahedron) in ascending order of face count. The bottom one shows solving a linguistic puzzle by selecting and placing the correct character block to complete two intersecting Chinese idioms.}
    \label{fig:reasoning_generation}
\end{figure}

Furthermore, as illustrated in Fig.~\ref{fig:reasoning_generation}, the system demonstrates reasoning abilities. These include geometric and relational inference for sorting tasks, as well as semantic structural reasoning for completing visual puzzles. Together, these capabilities push video generation beyond mere depiction toward dynamic, intelligent problem-solving.

\section{Conclusion}
In this report, we present Kling-Omni, a generalist generative model that bridges the traditional boundaries between video generation, editing, and multimodal reasoning. By leveraging a diffusion transformer aligned with a vision-language model, Kling-Omni establishes a shared embedding space that enables deep cross-modal interaction. Kling-Omni effectively replaces fragmented expert models with a single, holistic system capable of processing Multi-modal Visual Language (MVL) inputs to produce high-fidelity, physically plausible video content.

Our contributions extend beyond model architecture to encompass a robust training and data infrastructure. We constructed a comprehensive data engineering pipeline ensuring temporal stability and semantic alignment, and implemented a highly optimized infrastructure to ensure scalability and efficiency. Extensive evaluations demonstrate that Kling-Omni achieves state-of-the-art performance in complex tasks.

Looking forward, Kling-Omni represents a foundational step toward building multimodal world simulators capable of perceiving, reasoning, generating and interacting with the dynamic and complex worlds.

\section{Contributors}
All contributors are listed in alphabetical order by their last names.

Jialu Chen, Yuanzheng Ci, Xiangyu Du, Zipeng Feng, Kun Gai, Sainan Guo, Feng Han, Jingbin He, Kang He, Xiao Hu, Xiaohua Hu, Boyuan Jiang, Fangyuan Kong, Hang Li, Jie Li, Qingyu Li, Shen Li, Xiaohan Li, Yan Li, Jiajun Liang, Borui Liao, Yiqiao Liao, Weihong Lin, Quande Liu, Xiaokun Liu, Yilun Liu, Yuliang Liu, Shun Lu, Hangyu Mao, Yunyao Mao, Haodong Ouyang, Wenyu Qin, Wanqi Shi, Xiaoyu Shi, Lianghao Su, Haozhi Sun, Peiqin Sun, Pengfei Wan, Chao Wang, Chenyu Wang, Meng Wang, Qiulin Wang, Runqi Wang, Xintao Wang\footnote{Project Lead}, Xuebo Wang, Zekun Wang, Min Wei, Tiancheng Wen, Guohao Wu, Xiaoshi Wu, Zhenhua Wu, Da Xie, Yingtong Xiong, Yulong Xu, Sile Yang, Zikang Yang, Weicai Ye, Ziyang Yuan, Shenglong Zhang, Shuaiyu Zhang, Yuanxing Zhang, Yufan Zhang, Wenzheng Zhao, Ruiliang Zhou, Yan Zhou, Guosheng Zhu, Yongjie Zhu.

\newpage

\clearpage
\newpage
\bibliographystyle{kling/plainnat}
\bibliography{paper}

\clearpage
\newpage

\end{document}